\documentclass[10pt,journal,compsoc]{IEEEtran}

\usepackage{cite}
\usepackage{times}
\usepackage{epsfig}
\usepackage{graphicx}
\usepackage{amsmath}
\usepackage{amssymb}

\usepackage{gensymb}
\usepackage{dsfont}
\usepackage{wrapfig,lipsum,booktabs}
\usepackage{caption}
\usepackage{epsfig}
\usepackage{graphicx}
\usepackage{amsmath}
\usepackage{amsthm}
\usepackage{amssymb}
\usepackage{dsfont}
\usepackage{subfigure}
\usepackage{color}

\usepackage{afterpage}
\usepackage{float}
\usepackage[T1]{fontenc}
\usepackage{nopageno}
\usepackage{float}
\definecolor{mycolor}{RGB}{0,76,153}

\usepackage[pagebackref=true,breaklinks=true,colorlinks,bookmarks=false]{hyperref}

\begin{document}

\title{\Huge Self-supervised 3D Representation Learning of \\Dressed Humans from Social Media  Videos}

\author{Yasamin Jafarian \hspace{0.5in} Hyun Soo Park
\vspace{5pt}
\\
\textsuperscript{}{University of Minnesota} \\ {\tt\small \{yasamin, hspark\}@umn.edu}
\vspace{5pt}
}



\IEEEtitleabstractindextext{%

\begin{abstract}
A key challenge of learning a visual representation for the 3D high fidelity geometry of dressed humans lies in the limited availability of the ground truth data (e.g., 3D scanned models), which results in the performance degradation of 3D human reconstruction when applying to real-world imagery. We address this challenge by leveraging a new data resource: a number of social media dance videos that span diverse appearance, clothing styles, performances, and identities.
Each video depicts dynamic movements of the body and clothes of a single person while lacking the 3D ground truth geometry. To learn a visual representation from these videos, we present a new self-supervised learning method to use the local transformation that warps the predicted local geometry of the person from an image to that of another image at a different time instant. This allows self-supervision by enforcing a temporal coherence over the predictions.
In addition, we jointly learn the depths along with the surface normals that are highly responsive to local texture, wrinkle, and shade by maximizing their geometric consistency.
Our method is end-to-end trainable, resulting in high fidelity depth estimation that predicts fine geometry faithful to the input real image. 
We further provide a theoretical bound of self-supervised learning via an uncertainty analysis that characterizes the performance of the self-supervised learning without training.
We demonstrate that our method outperforms the state-of-the-art human depth estimation and human shape recovery approaches on both real and rendered images.
\end{abstract}

}

\maketitle



\section{Introduction}

\begin{figure}[t]
  \begin{center}
    \includegraphics[width=0.48\textwidth]{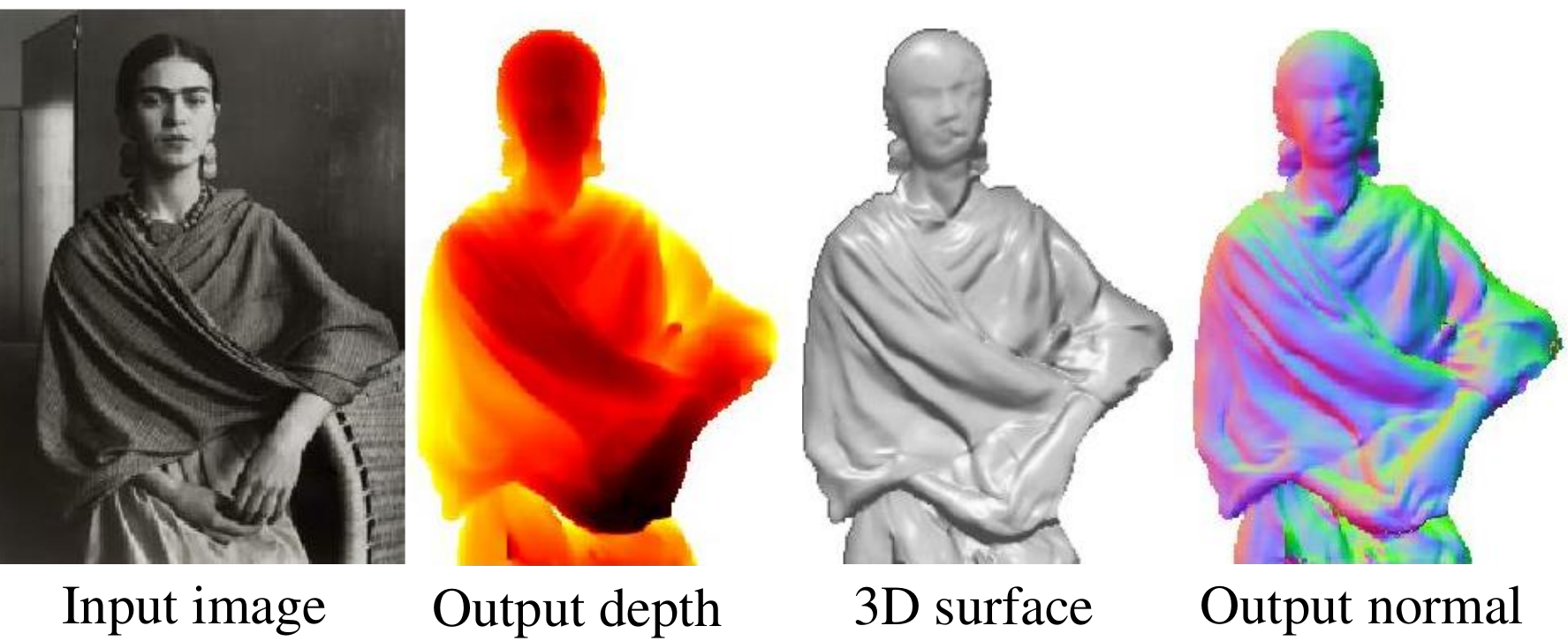}
  \end{center}
     \vspace{-4mm}
  \caption{We present a novel method that takes as input an image of dressed human(s) and outputs high
fidelity depths and its surface normals. The estimated depths capture fine wrinkles of scarf, dress,
and body shape. Photograph of Frida Kahlo by Imogen Cunningham.}
  \label{Fig:frida}
\end{figure}

Consider a historic photograph of Frida Kahlo wearing a beautiful shoulder scarf and ornaments as shown in Figure \ref{Fig:frida}. We as humans can effortlessly perceive the fine-grained 3D geometry of her face, draped scarf, hair, and earrings from this 2D photograph. Can a machine be equipped with such perceptual capability such that we can travel back to the early 1930s to see her lively moment? With the increasing prevalence of VR and AR, this perceptual capability to precisely model the complex geometry of humans is becoming the key to authentic social tele-presence. 
Note that existing parametric body models for humans such as SMPL and its variants~\cite{SMPL:2015, SMPL-X:2019, Bogo:2016, Lassner:2017,Moon_2020_ECCV_I2L, Choi_2020_ECCV_Pose2Mesh, choi2020beyond, Rhodin:2016, Kanazawa:2018, Omran:2018,xiang2019monocular} have limited expressibility to model the complex 3D geometry of dressed humans. 

To capture the fine-grained 3D geometry, e.g., wrinkle and fabric texture, photogrammetry based on massive camera infrastructure (e.g., 40-500 cameras to cover full body shape)~\cite{joo_cvpr_2014,Bee10,Wenger:2005} has been used, resulting in production-level rendering~\cite{LOMBARDI:2018,armando:2018,tao2021function4d} and 3D fabrication~\cite{twindom,shapify}. Despite its promise, the practical deployment of such massive camera systems in our daily environment is still challenging because of its hardware requirements and computational complexity. 
Single view reconstruction is an immediate remedy to address this challenge where 3D representation of humans can be learned in a supervised fashion from the scanned human 3D models ~\cite{RP:2020,axyz:2020,twindom,tao2021function4d}. Nonetheless, the number of these 3D data to train such model is limited (e.g., a few hundreds of static models), which do not span diverse poses, appearance, and complex cloth geometry resulting in the performance degradation of 3D human reconstruction when applying to real-world imagery. This makes a sharp contrast with the existing datasets available for scene understanding (e.g., ScanNet~\cite{dai2017scannet}) that are made of millions of data instances to learn geometry and visual semantics, i.e., the size of data for humans is at least one or two orders of magnitude smaller than that for scenes.

\begin{figure*}[!ht]
\begin{center}
\includegraphics[width=1\textwidth]{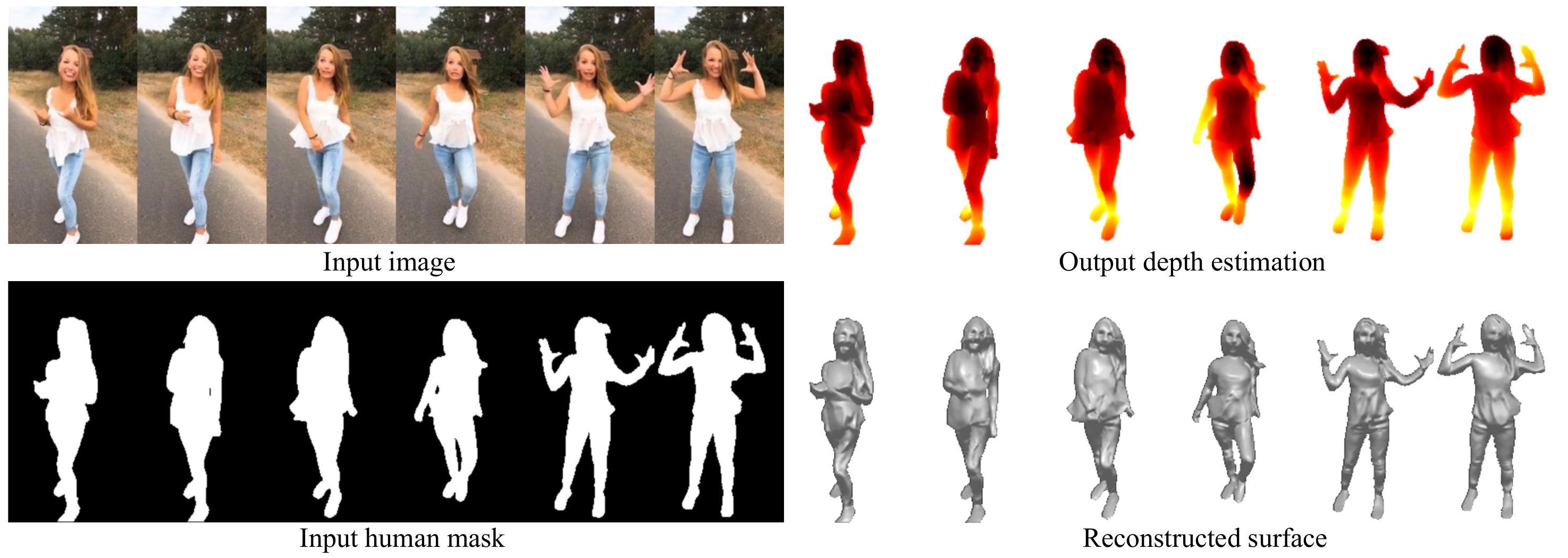}
\end{center}
    \vspace{-3mm}
   \caption{This paper presents a novel approach to estimate high fidelity depths of dressed humans from a single view image by leveraging a new data resource: a number of social media dance videos that span diverse appearance, clothing styles, performances, and identities. We show an example sequence and the corresponding human mask along with the estimated depth (the darker, the closer) and the reconstructed surface.}
\label{fig:teaser}
\end{figure*}

In this paper, we fill this data gap by utilizing a new type of human visual data---hundreds of dance videos shared in social media (e.g., TikTok mobile application) that span diverse appearance, pose, shape, and identities. This enables us to reconstruct high fidelity 3D geometry of dressed humans in the form of depths and surface normals from a single view image as shown in Figure~\ref{fig:teaser}. The main characteristics of these dance videos are that 1) each video depicts a sequence of diverse poses of a single person; and 2) 3D ground truth is not available, i.e., existing fully supervised paradigm is not applicable. 

To learn a visual representation of high fidelity 3D human geometry from social media dance videos, we make use of two geometric properties agnostic to 3D ground truth. (1) Local shape invariance: we conjecture that since the geometry of dressed humans is inherently semi-rigid so that the local geometry of the same person approximately remains constant up to some pose transformations. For instance, the cloth movement on the left upper arm region undergoes approximately a rigid transformation when its pose changes. Therefore, it is possible that the geometric consistency over different poses can be applied to learn from the real dance videos. We estimate a transformation for each body part that can warp its 3D geometry from one image to another image at a different time instant. This allows us to self-supervise the predicted geometry of the dressed humans without 3D supervision. 
This geometry transformation allows further applying photometric consistency.
 (2) Geometric consistency: while modern learning based depth estimators are capable of recovering holistic scene geometry, it often fails to 
encode fine local geometry such as complex cloth wrinkles and face profile features~\cite{mannequin:2019}, which constitutes the dominant factor of realism. On the other hand, surface normals are highly responsive to fine visual structures such as texture and wrinkles~\cite{wang2020normalgan}. We exploit the geometric relationship to jointly learn depths and surface normals (e.g. matching the surface normal to the curvature of the depth). 

Our end-to-end trainable method takes as input an RGB image, the corresponding human foreground, and human UV coordinates and outputs high fidelity depths of fine wrinkles and shapes that are faithful to the input image. We design a network called \textit{HDNet} that learns to predict the depths and surface normals, and the predicted surface normals are, in turn, used to ensure the geometric consistency with the predicted depths. We use a Siamese design of HDNet to measure the self-consistency across time by warping one prediction to another. To the end, our method is semi-supervised by leveraging both 3D scanned models and real dance videos.  We demonstrate that our method outperforms the state-of-the-art human depth estimation approaches on both real and rendered images.

Our core contributions include: 
(1) a new dataset called \textit{TikTok dataset} that consists of more than 340 sequences of dance videos shared in a social media mobile platform, TikTok, totaling more than 100K images along with the human mask and human UV coordinates; (2) a novel formulation that warps the 3D geometry of dressed humans from one image to the other image at a different time instant to measure self-consistency, which allows us to utilize the real dance videos; (3) HDNet design that learns to predict fine depths reflective of surface normal prediction by enforcing their geometric consistency; (4) strong qualitative and quantitative prediction on real world imagery. 

Building upon the earlier version \cite{Jafarian_2021_CVPR_TikTok}, we make the following additional contributions:
(1) incorporating photometric consistency to learn the visual representation; (2) generalizing Euclidean transformation to affine transformation to handle  large deformation (3) including a new baseline PaMIR \cite{zheng2020pamir} and a new evaluation on THuman2.0 \cite{tao2021function4d} dataset.

\section{Related Works}

Our paper tackles a problem that lies at the intersection of human body reconstruction, single view depth estimation, and human 3D datasets.

\noindent\textbf{Human Body Reconstruction} There are two predominant representations in human body reconstruction: parametric and non-parametric. Similar to face modeling~\cite{cootes:2001}, parametric mesh models such as SCAPE~\cite{SCAPE:2005} and SMPL~\cite{SMPL:2015} are an attractive choice of the human body representation, which can be used for single view human reconstruction~\cite{Bogo:2016, Rhodin:2016, Kanazawa:2018, Lassner:2017, Omran:2018,SMPL-X:2019,xiang2019monocular, Moon_2020_ECCV_I2L, Choi_2020_ECCV_Pose2Mesh, choi2020beyond} and synthetic data generation~\cite{Varol:2017, Varol:2018}. 
The number of parameters to model a 3D full body is relatively small (pose and shape), which makes regressing the parameters from a single view image possible. 
However, despite their remarkable performance, the reconstructed geometry has a limited resolution predefined by the mesh topology, which prevents them from expressing the fine details of dressed humans. 
For example, the fine-grained 3D geometry such as clothes and hair cannot be modeled. 
These challenges have been addressed by refining parametric models with residual geometry~\cite{Alldieck:2018,alldieck2019tex2shape,ma20autoenclother,Laehner:2020}. Depth~\cite{Deephuman:2019, mannequin:2019} or volumetric representation~\cite{deepcap,Zheng2019DeepHuman} as a non-parametric representation can describe the geometry of dressed humans. 
Tan et al. \cite{tan2020self} combined the parametric and nonparametric representations in a semi-supervised manner by leveraging the videos of people in motion.
They predicted the SMPL representation and then refined it to include more details on the surface by leveraging the photometric consistency in the temporal domain. However, the complex clothing items such as skirts still cannot be captured in this method. 
A non-parametric representation is a viable solution to model such complex geometry of dressed humans. However, it requires a large amount of 3D ground truth data to predict a number of parameters (e.g., for depths, the number of prediction variables is comparable to the number of pixels).  Li et al.~\cite{mannequin:2019} addressed this challenge by exploiting a large community dataset of Mannequin Challenge, and Tang et al.~\cite{Deephuman:2019} incorporated semantic labels (pose and segmentation) to regularize their depth estimator.

\noindent\textbf{Single View Depth Estimation} Single view depth estimation is a core task of scene understanding where sophisticated designs of convolutional neural networks (CNNs) enable predicting scene geometry \cite{Luo-VideoDepth-2020}. To capture fine details of depth reconstruction, additional cues such as surface normals have been incorporated~\cite{zhang2018deepdepth, GeoNet:2018, Deephuman:2019, Nehab:2005, Qiu_2019_CVPR, feiWS19, Or-el2015, SimPose2020}. Iterative least squares \cite{Deephuman:2019} and kernel regression \cite{GeoNet:2018} have been used to fuse the surface normals and depths, and coarse-to-fine learning is used to densify LiDAR data for outdoor scenes or missing depth data~\cite{Qiu_2019_CVPR} for indoor scenes~\cite{zhang2018deepdepth}. 
Recently, integrating the surface normal into the depth prediction \cite{wang2020normalgan} (e.g. identifying whether a normal representation is realistic or not using GAN \cite{GAN}) has shown to be effective in restoring local geometry such as cloth wrinkles and face profile features. Unlike previous work, we focus on recovering sub-centimeter detailed geometry tailored to dressed humans by jointly learning depths and surface normals and leveraging a large dataset of social media dance videos. Unfortunately, to date, there exists no human visual data of which scale is comparable to the scene understanding datasets such as ScanNet~\cite{dai2017scannet} and KITTI~\cite{Geiger2012CVPR,Geiger2013IJRR, Fritsch2013ITSC, Menze2015CVPR}. This presents a new challenge for learning a visual representation for human single view depth estimation. 

\noindent\textbf{Human 3D Datasets} While there are a number of RGBD 
datasets for structural scene understanding \cite{Matterport3D, dai2017scannet, Xiao:2013:SAD, silberman:2012}, a limited amount of data address the problem of the geometry prediction for dressed humans
 in the wild. A few RGBD datasets~\cite{Shahroudy_2016_NTURGBD , Liu_2019_NTURGBD120 , UTD-MHAD , Barbosa:reid12} are designed for humans action recognition. However, these data lack the  geometric details such as cloth wrinkles. For human geometry, the 3D scanned models \cite{RP:2020 , axyz:2020 , twindom} or multiview generated models \cite{adobe_data:2008, buff} can be used to generate photorealistic images from multiple views, which has been used for training a geometry predictor with full supervision~\cite{pifu:2019, pifuhd:2020}. However, the amount of data is still limited to a few hundreds of static models, which prevents learning a model that can predict images of humans in the wild. In this paper, we introduce a new source of data: real dance videos from social media to generalize the human depth estimation to different viewpoints, human appearance, clothing styles and poses.


\section{Method}

Given a single image of a dressed human $\mathbf{I}$, we reconstruct its high fidelity depth, i.e., $z = g(\mathbf{x};\mathbf{I})$,
where $\mathbf{x} \in \mathds{R}^{2}$ is the $xy$-location in the image, and $z \in \mathds{R}_+$ is the depth at the corresponding location.

Existing approaches learn $g$ directly from the ground truth data, which shows 
two limitations in estimating depths of dressed humans. (1) While existing depth estimators are highly responsive to predict  holistic scene geometry, it is shown~\cite{mannequin:2019} that its expressibility is limited at 
encoding fine local geometry such as irregular and complex wrinkles, which constitute the dominant factor of human geometry/rendering realism.  (2) It requires a large amount of 3D ground truth data (e.g., ScanNet~\cite{dai2017scannet} and KITTI~\cite{Geiger2012CVPR,Geiger2013IJRR}). Such large ground truth data for humans that span diverse appearance, cloth styles, and poses do not exist (e.g., a few hundreds of posed scanned models~\cite{RP:2020,axyz:2020,twindom}).

\subsection{Self-supervised Human Depths from Videos} 
We present a new method to address these limitations by leveraging large video data of real humans in motion. Albeit lacking of 3D ground truth, each video depicts the movement of a single person across time where her/his geometry approximately remains constant up to local transformations.

\begin{figure}[t]
  \begin{center}
  \vspace{-3mm}
    \includegraphics[width=0.475\textwidth]{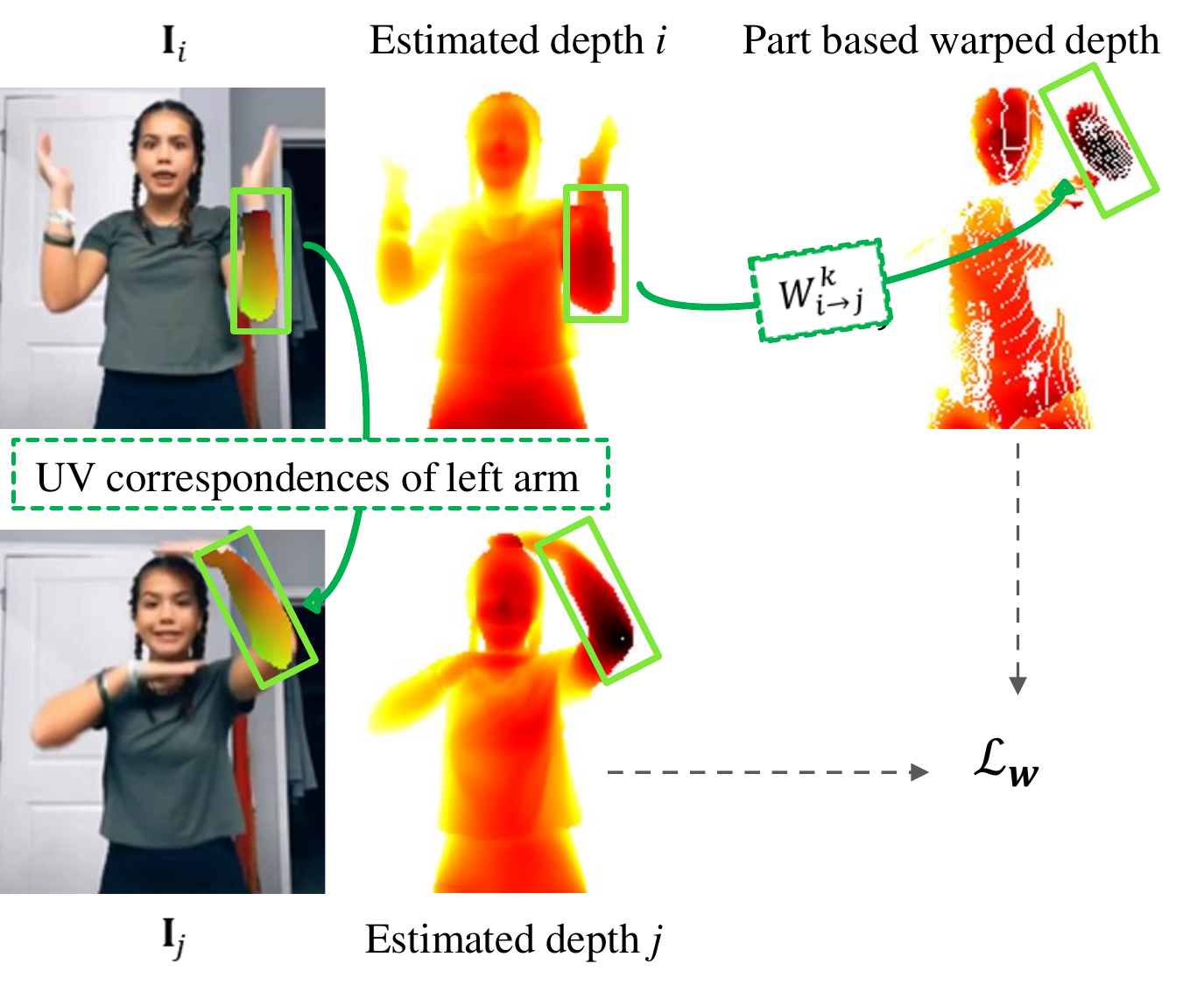}
  \end{center}
     \vspace{-6mm}
  \caption{ Given the depth estimate at the $i^{\rm th}$ time instant, we use a part based transformation that warps the 3D local geometry of the image to the image at the $j^{\rm th}$ time instant. The green boxes in two images show the UV correspondences of the left arm. The depths of the left arm are reconstructed in 3D and transformed to the $j^{\rm th}$ time to form the part based warped depths to supervise the depth estimate at the $j^{\rm th}$ time instant through the warping loss $\mathcal{L}_w$.}
  \label{Fig:warping_sum}
\end{figure}

Consider a coordinate transform $h(\mathbf{u}) = \mathbf{x}$ that maps a canonical human body surface coordinate $\mathbf{u}\in\mathds{R}^2$ (UV surface coordinate) to the corresponding point $\mathbf{x}$ in an image. A key feature of the UV surface coordinate is that it is invariant to poses, clothes, and appearance.

We parametrize a 3D point $\mathbf{p}\in \mathds{R}^3$ reconstructed by the depth prediction using the UV coordinate, i.e., 
\begin{align}
    \mathbf{p}_i(\mathbf{u}) = z \mathbf{K}^{-1}\widetilde{\mathbf{x}} = g(h_i(\mathbf{u}); \mathbf{I}_i) \mathbf{K}^{-1} \widetilde{h}_i(\mathbf{u}),
     \label{Eq:3d_project}
\end{align}
where $\mathbf{K}\in \mathds{R}^{3\times3}$ is the camera intrinsic parameter, $\widetilde{{\cdot}}\in \mathds{P}^2$ is the homogeneous representation~\cite{hartley:2004}, and $\mathbf{x}$ is the pixel location in the image domain corresponding to $\mathbf{u}$ in the UV domain. The subscript $i$ indicates the time instant. 

We transform a set of points in the $k^{\rm th}$ body part at the $i^{\rm th}$ time instant to the $j^{\rm th}$ time instant:
\begin{align}
    \mathbf{p}_{i\rightarrow j}(\mathbf{u}) = \mathcal{W}_{i\rightarrow j}^k(\mathbf{p}_i(\mathbf{u})), ~~~~\mathbf{u} \in \mathcal{U}_k 
     \label{Eq:rigid_transform}
\end{align}
where $\mathcal{W}$ is a 3D part based warping function, and $\mathcal{U}_k$ is the set of UV coordinates associated with the $k^{\rm th}$ body part. The body part is defined as a region of the body where its local geometry approximately undergoes a parametric 3D transformation such as affine or rigid, e.g., lower arm. An analogous warping is used for non-rigid tracking~\cite{Newcombe:2015} without the part based representation, which allows mapping between consecutive frames. With the part based warping, we substantially extend the time horizon by parametrizing the 3D point using the UV coordinate, which does not require an offline iterative closest point method between the consecutive frames. 
 
We use an affine transformation, i.e., $\mathcal{W}_{i\rightarrow j}^k(\mathbf{p}_i) = \mathbf{A}^k_{i\rightarrow j}\mathbf{p}_i+\mathbf{t}^k_{i\rightarrow j}$ where $\mathbf{A}$ is a $3\times3$ nonsingular matrix, and $\mathbf{t}$ is a $3\times1$ translational vector.
With the pre-defined correspondences, we compute the transformation by minimizing the following error:
\begin{align}
    \underset{\mathbf{A},\mathbf{t}}{\operatorname{minimize}} \sum_l \left\|\mathbf{p}_j(\mathbf{v}_l)-\mathcal{W}^k_{i\rightarrow j}(\mathbf{p}_i(\mathbf{v}_l))\right\|^2,~~ \mathbf{v}_l\in \mathcal{V}_k \subset \mathcal{U}_k, \nonumber
\end{align}
where $\mathcal{V}_k$ is the subset of the UV coordinates that represent the overall transformation. We minimize the objective using least squares~\cite{Arun:1987}. In practice, we choose the sparse correspondences in the subset by discretizing the UV coordinates. This transformation is computed online, i.e., the transformation changes as the depth prediction is updated at each training iteration.

Figure \ref{Fig:warping_sum} illustrates the self-supervision via warping the 3D geometry of humans between two arbitrary frames of a video. We use the UV coordinates to warp the estimated depth for each body part from the $i^{\rm th}$ time instant to the $j^{\rm th}$ time instant, resulting in a sparse warped depth that can supervise the depth estimate at the $j^{\rm th}$ time instant by minimizing warping loss $\mathcal{L}_w$.

We minimize the following loss to measure geometric discrepancy between two time instances:
\begin{align}
    \mathcal{L}_w = \sum_{l}\sum_{(i,j)\in\mathcal{V}_l} \sum_k \sum_{\mathbf{u}\in\mathcal{U}_k} \|\mathbf{p}_j (\mathbf{u}) - \mathbf{p}_{i\rightarrow j}(\mathbf{u})\|^2,
    \label{Eq:temporal_loss_eq_uv} 
\end{align}
where $\mathcal{V}_l$ is the set of time instances within the $l^{\rm th}$ video. 

With the warped geometry, we can further enforce photometric consistency, i.e.,
\begin{align}
    \mathcal{L}_p = \sum_{l}\sum_{(i,j)\in\mathcal{V}_l}\sum_k\sum_{\mathbf{u}\in\mathcal{U}_k} \|\mathbf{\mathbf{I}}_j (\mathbf{x}_{i\rightarrow j}(\mathbf{u})) - \mathbf{\mathbf{I}}_{i}({h}_{i}(\mathbf{u}))\|^2,
    \label{Eq:temporal_loss_photo} 
\end{align}
where $\widetilde{\mathbf{x}}_{i\rightarrow j}(\mathbf{u}) = \lambda \mathbf{K}\mathbf{p}_{i\rightarrow j}(\mathbf{u}) $  is the 2D projection of $\mathbf{p}_{i\rightarrow j}(\mathbf{u})$ to the $j^{\rm th}$ time instant.

Equation (\ref{Eq:temporal_loss_eq_uv}) and (\ref{Eq:temporal_loss_photo}) allows us to utilize a large amount of real videos without the 3D ground truth via self-supervision, i.e., the estimated depth in one pose can be used to supervise the depth in the other pose. This makes the depth estimation responsive to real data of diverse human poses and appearances.

\subsection{Joint Learning of Surface Normal and Depth}
Surface normals are known to be highly correlated with the local texture, wrinkle, and shade~\cite{wang2020normalgan,Deephuman:2019} 
because of its first order nature of pixel intensity, i.e., under Lambertian lighting model, the pixel intensity is linear in the surface normal.
We jointly estimate surface normals and depths to benefit from each other. We estimate the surface normals of an image $\mathbf{I}$, i.e., $\mathbf{n} = f(\mathbf{x};\mathbf{I})$
where 
$\mathbf{n} \in \mathds{S}^2$ is the unit surface normal vector represented in the camera coordinate system.

Surface normal $\widehat{\mathbf{n}}(\mathbf{x})$ is the curvature that is perpendicular to the tangential plane of the corresponding 3D point $\mathbf{p(x)}$ (we override the notation $\mathbf{p}(\mathbf{u})$ in Equation (\ref{Eq:3d_project})), i.e., 
\begin{align}
    \widehat{\mathbf{n}}(\mathbf{x}) = \frac{\partial \mathbf{p}({\mathbf{x}})}{\partial x} \times \frac{\partial \mathbf{p}({\mathbf{x}})}{\partial y} / \left\|\frac{\partial \mathbf{p}({\mathbf{x}})}{\partial x} \times \frac{\partial \mathbf{p}({\mathbf{x}})}{\partial y}\right\|,
    \label{Eq:pde}
\end{align}
where $\widehat{\mathbf{n}}$ denotes the surface normal estimate derived by the depth estimate.

We ensure geometric consistency between the predicted surface normals and the derived surface normals from the depth estimates by minimizing their geometric error:
\begin{align}
    \mathcal{L}_s =  \sum_{\mathbf{I}_i\in \mathcal{D}}\sum_{\mathbf{x} \in \mathcal{R}(\mathbf{I}_i)} \cos^{-1}\left(\frac{\mathbf{n}^\mathsf{T}(\mathbf{x}) \widehat{\mathbf{n}}(\mathbf{x})}{\|\mathbf{n}(\mathbf{x})\|\|\widehat{\mathbf{n}}(\mathbf{x})\|}\right), \label{Eq:loss_geo}
\end{align}
where $\mathcal{R}(\mathbf{I})$ is the coordinate range of the image $\mathbf{I}$, and $\mathcal{D}$ is the image dataset including the dance videos and scanned 3D models.

\begin{figure}[t]
  \begin{center}
    \includegraphics[width=0.47\textwidth]{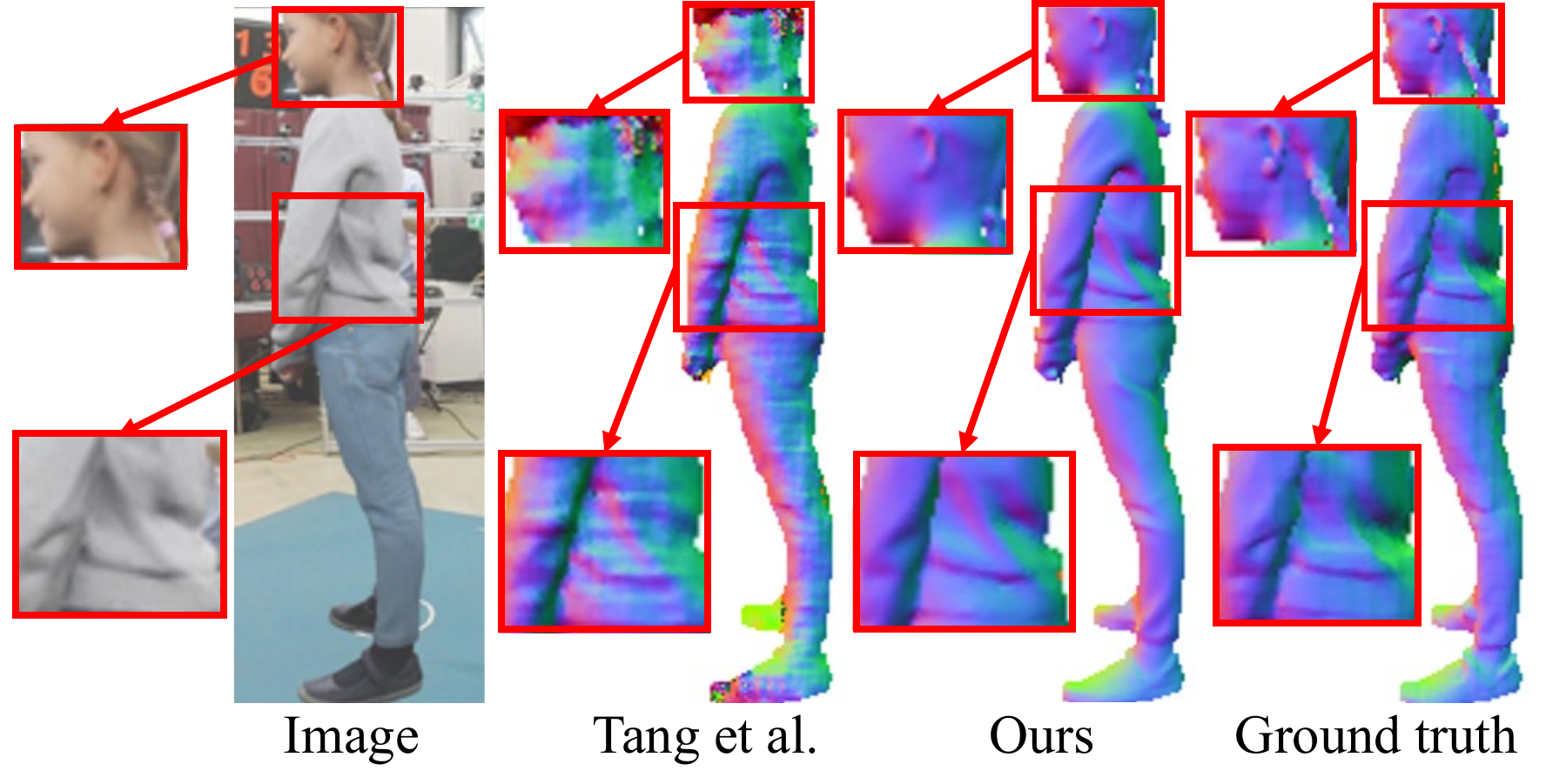}
  \end{center}
     \vspace{-6mm}
  \caption{We compare our method with Tang et al.~\cite{Deephuman:2019} on the surface normals derived from the depths. While two methods use the surface normals to enhance the depths, unlike Tang et al., our method jointly learns surface normals and depths by supervising them with each other, which produces more realistic and less noisy prediction that preserves the detailed geometry of wrinkles and face.  
}
  \label{Fig:tangvsours}
\end{figure}

Note that the relationship between surface normal and depth has been used to obtain the details of depth estimates. GeoNet~\cite{GeoNet:2018} has leveraged the derived surface normals from the predicted depths to refine the predicted surface normals for an indoor scene understanding. In human domain, Tang et al.~\cite{Deephuman:2019} uses the surface normal prediction to refine the human depth prediction in a post-processing manner. Unlike these methods, we use the surface normal estimates to supervise the depths and the depth estimates to supervise the surface normals by enforcing their geometric consistency in the training phase. This end-to-end online pipeline enables learning the depths from the real videos without the ground truth depth. Figure \ref{Fig:tangvsours} illustrates the comparison of the surface normal generated from the predicted depth of our method and Tang et al. \cite{Deephuman:2019}. Our result is realistic, which captures the wrinkles of the cloth fabric compared to Tang et al. \cite{Deephuman:2019}.

\begin{figure*}[th]
\begin{center}
\subfigure[Network design of HDNet.]
    {\label{Fig:net}\includegraphics[width=1\textwidth]{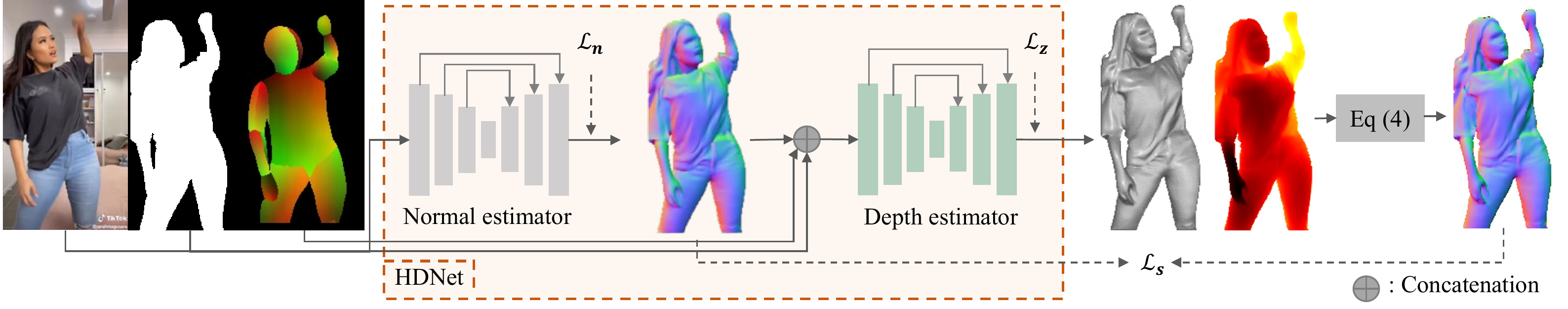}}\\\vspace{-2mm}
\subfigure[HDNet self-supervision using two images from different time instances.]
    {\label{Fig:warp}\includegraphics[width=1\textwidth]{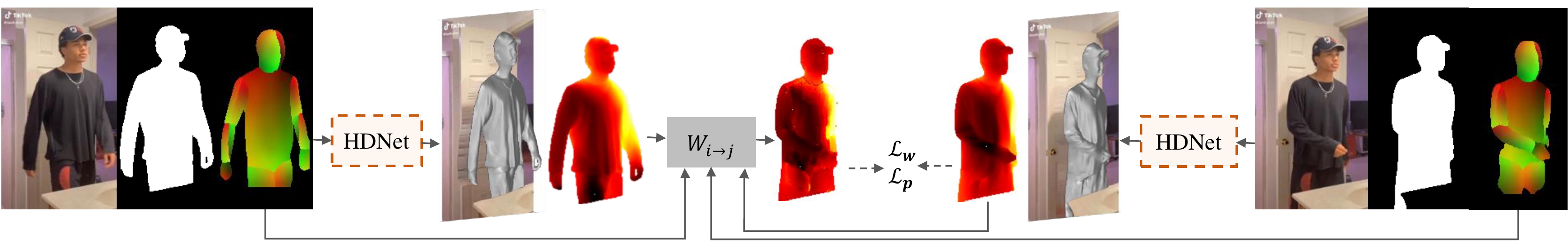}}\vspace{-2mm}
\end{center}
\vspace{-3mm}
   \caption{(a) Our network \textit{HDNet} takes as input an image with the correspondending human foreground and UV coordinates and predicts the high fidelity depths of the dressed human.  The HDNet is composed of the depth and surface normal estimators. The surface normal estimator takes as input, an image and its foreground human mask and outputs the surface normals. The estimated surface normals are, in turn, used as an input along with the image, foreground human mask, and part based UV coordinate to the depth estimator. We enforce the geometric consistency between the estimated depths and surface normals. (b) We build a Siamese design of HDNet to leverage real dance videos. The estimated depth of one image is warped to the other image at a different time instant using a part based transformation. We measure the geometric and photometric consistency between the predicted depths and warped depths through $\mathcal{L}_w$ and $\mathcal{L}_p$ respectively.}
\label{Fig:network}
\end{figure*}

\begin{figure*}[t]
  \begin{center}
    \includegraphics[width=1\textwidth]{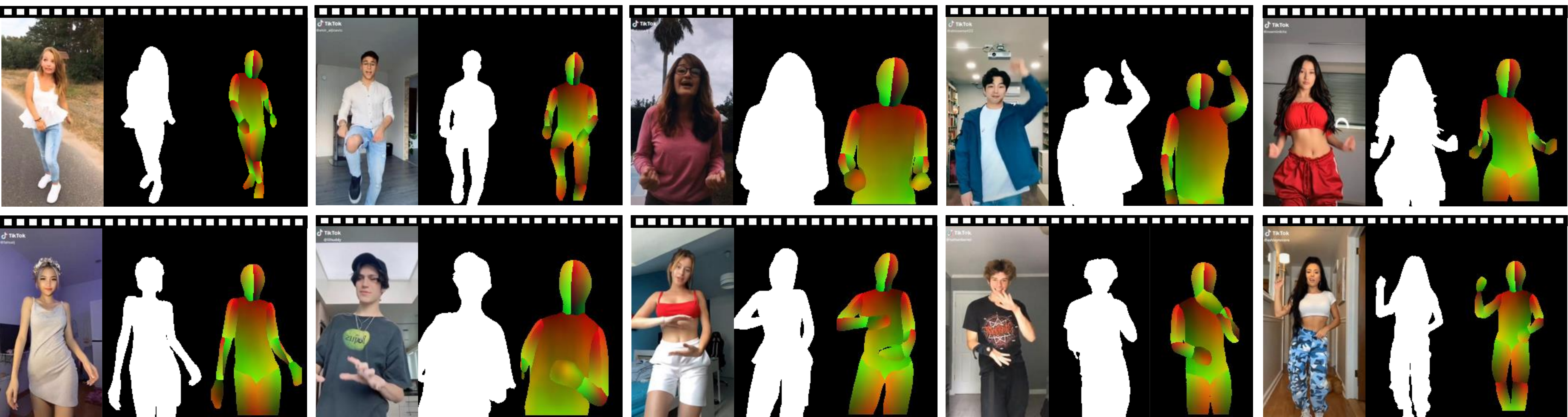}
  \end{center}
     \vspace{-3mm}
  \caption{TikTok Dataset. We present a new dataset called \textit{TikTok dataset} that consists of 340 sequences of dance videos shared in a social media mobile platform, TikTok, totaling more than 100K images along with the human mask and human UV coordinates.}
  \label{Fig:dataset}
\end{figure*}

\subsection{Network Design}

We minimize the following overall loss to learn the depth and surface normal estimators from real videos and 3D scanned models:
\begin{align}
    \mathcal{L}  = 
    \mathcal{L}_{z}+ \lambda_{n} \mathcal{L}_{n} + \lambda_{s} \mathcal{L}_{s} + \lambda_{w} \mathcal{L}_{w} + \lambda_{p} \mathcal{L}_p,
\end{align}
where $\lambda_{n}$, $\lambda_{s}$, $\lambda_{w}$, and $\lambda_{p}$ are relative weights between losses. 
In addition to self-consistency losses ($\lambda_{w}$ and $\lambda_{s}$), we utilize the 3D ground truth data from the 3D scanned models~\cite{RP:2020}.
This depth and surface normal can be learned by minimizing the following error between ground truth normal ${\mathbf{N}}(\mathbf{x})$ and the prediction.
\begin{align}
    \mathcal{L}_{z} = \sum_{\mathbf{I}\in \mathcal{D}_s}\sum_{\mathbf{x} \in \mathcal{R}(\mathbf{I})}\|{{Z}}(\mathbf{x})-g(\mathbf{x};\mathbf{I})\|^2, 
\end{align}
\begin{align}
    \mathcal{L}_n =  \sum_{\mathbf{I}\in \mathcal{D}_s}\sum_{\mathbf{x} \in \mathcal{R}(\mathbf{I})} \cos^{-1}\left(\frac{\mathbf{N}^\mathsf{T}(\mathbf{x}) f(\mathbf{x};\mathbf{I})}{\|\mathbf{N}(\mathbf{x})\|\|f(\mathbf{x};\mathbf{I})\|}\right), \label{Eq:loss_all}
\end{align}
where $\mathcal{D}_s$ is the 3D scanned dataset with the ground truth depths ${{Z}}(\mathbf{x})$ and surface normals ${\mathbf{N}}(\mathbf{x})$.

\noindent\textbf{Network Design and Details} 
We design our neural network called \textit{HDNet} (Human Depth Neural Network) that allows us to utilize both real videos and 3D scanned model data as shown in Figure~\ref{Fig:net}. HDNet is composed of two estimators: surface normal and depth estimators. The surface normal estimator $f(\mathbf{x};\mathbf{I})$ takes as input an RGB image and its foreground mask, and outputs the surface normal estimates. The depth estimator, $g(\mathbf{x};\mathbf{I})$, in turn, takes as input a triplet of an RGB image, foreground mask, and UV coordinate, and outputs the depth estimates. The geometric consistency between the surface normal and depth is enforced by minimizing $\mathcal{L}_s$. For the 3D scanned model data, both estimators are supervised by the ground truth surface normal and depth ($\mathcal{L}_n$ and $\mathcal{L}_z$), respectively. 

For the real videos, we build a Siamese network with HDNet where two triplets from two time instances within the same video are used for the depth estimates as shown in Figure \ref{Fig:warp}. The UV coordinates from both images are used to compute the affine transformation that is used to warp the depth from one image to the other image. At each time instant, we make five image pairs by randomly selecting the time instances that are at least 5 frames apart.

For the two estimators, we use the stacked hourglasses network~\cite{newell:2006} as a backbone network. The image and its foreground mask are cropped from the input image and resized to 256$\times$256, and $h$ is approximated by the inverse of the UV map obtained by DensePose~\cite{Guler2018DensePose}. 
We use Adam optimizer~\cite{kingma:2015} with the following parameters for the training. Batch size: 10; learning rate: 0.001; the number of epochs: 380; $\lambda_\mathbf{n}$: 1; $\lambda_\mathbf{s}$: 0.5; $\lambda_\mathbf{w}$: 5, and $\lambda_\mathbf{p}$: 5; GPU model: NVIDIA V100. 

\begin{figure*}[t]
    \begin{center}
        \includegraphics[width=0.99\textwidth]{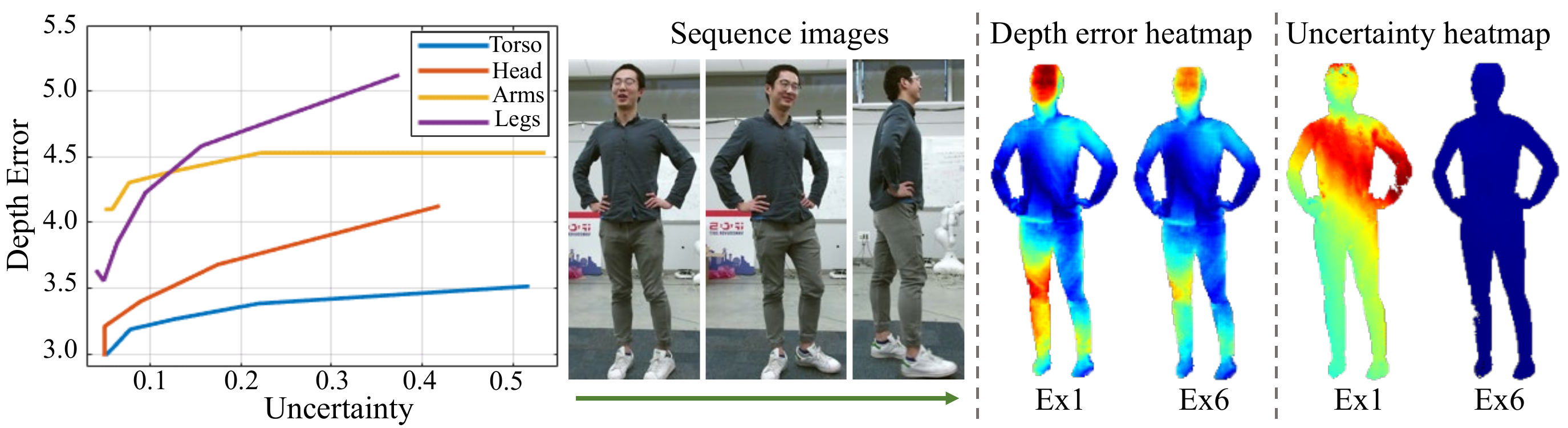}
    \end{center}
    \vspace{-4mm}
    \caption{The theoretical bound of depth prediction with respect to uncertainty on Tang et al. dataset \cite{Deephuman:2019}. From left to right, we show (1) the depth prediction error as a function of reconstruction uncertainty on torso, head, arms, and legs area, which shows that the performance of the depth prediction by self-supervision is bounded by the uncertainty, (2) the first, middle, and last frames of the sequence, (3) depth error heatmap for Ex1 (small motion) and Ex6 (large motion), and (4) the uncertainty heatmap for Ex1 and Ex6.}
    \label{Fig:unceVsErr}
    \vspace{-4mm}
\end{figure*}

\noindent\textbf{Noisy UV Filtering}  For every pair of images with the DensePose correspondence, we evaluate the validity of correspondences based on two criteria. (1) The pair must share at least five visible body parts where each common body part is defined by the one with more than 50 overlapping UV correspondences. (2) We choose the pairs to be at least 5 frames apart to ensure sufficient motion between frames by assuming that the noise in the prediction is not coherent across time. For minor spurious correspondences, (i.e. some noisy pixels in a specific body part) it is handled by the least squares warping solving.

\section{TikTok Dataset} \label{Sec:dataset}

We learn high fidelity human depths by leveraging a collection of social media dance videos scraped from the TikTok mobile social networking application. It is by far one of the most popular video sharing applications across generations, which include short videos (10-15 seconds) of diverse dance challenges as shown in Figure~\ref{Fig:dataset}. We manually find more than 300 dance videos that capture 
a single person performing dance moves from TikTok dance challenge compilations for each month, variety, type of dances, which are moderate movements that do not generate excessive motion blur.   
For each video, we extract RGB images at 30 frame per second, resulting in more than 100K images. We segmented these images~\cite{removebg:2020}, and computed the UV coordinates. The dataset and code can be found in \url{https://www.yasamin.page/hdnet_tiktok}.

\section{Theoretical Analysis on Self-supervised Learning via Reconstruction Uncertainty}

We formulate our self-supervised learning based on the assumption that the depth predictions in different time instances can be complementary to each other, i.e., a depth prediction in one frame can provide a new information to that of another frame. This assumption applies a majority of our dance videos while there are a few trivial cases where the assumption does not apply effectively. One extreme case would be a sequence of static pose (no motion). Since the depth prediction from one frame does not provide any new information to other frames, the self-supervised learning must be not effective. We characterize the impact of the body motion with respect to the depth prediction using an uncertainty analysis. This analysis allows us to anticipate the impact of the self-supervised learning without training as it provides theoretical performance bound of the self-supervised learning.\\
    \noindent\textbf{Uncertainty Modeling:} 
    Consider a 3D point $\mathbf{X}_i\in\mathds{R}^3$ reconstructed by the $i^{\rm th}$ time instant. The uncertainty of the point can be modeled by a covariance matrix $\mathbf{C}_\mathbf{X}\in \mathds{R}^{3\times3}$ where the covariance specifies the direction of uncertainty, i.e., the singular vectors and values. The 3D point is visible in the $j^{\rm th}$ time instant, which can be mapped to the $i^{\rm th}$ time instance to form $\mathbf{X}_{j\rightarrow i}$. This allows us to find the expected value and its covariance:
     \begin{align}\label{eq:Enfunc}
    \mathbf{C}_{\textbf{X}} = \mathds{E}[(\textbf{X}_{i} - \mathds{E}[\textbf{X}_{i}]) (\textbf{X}_{i} - \mathds{E}[\textbf{X}_{i}])^{\mathsf T}], \\~~~~ {\rm where}~~~~
    \mathds{E}[\textbf{X}_i] = \frac{1}{T} \sum_{j=1}^T {\textbf{X}}_{j\rightarrow i},
    \end{align}
    where $\mathds{E}[\mathbf{X}]$ is the expected value of $\mathbf{X}$. The covariance of the 3D point can be measured by:
    \begin{align}\label{eq:Enfunc}
    u(\mathbf{X}) = \sum_k \sigma_k, ~~~~\\{\rm where}~~~\mathbf{C}_{\mathbf{X}} = \mathbf{U} \boldsymbol{\Sigma} \mathbf{V}^{\mathsf T} = \mathbf{U} \begin{bmatrix}
                 \sigma_{1}&0&0\\
                 0&\sigma_{2}&0\\
                 0&0&\sigma_{3}
                 \end{bmatrix} \mathbf{V}^{\mathsf T},
    \end{align}
    where $u(\mathbf{X})$ is the uncertainty of $\mathbf{X}$ that is a sum of singular values of the covariance matrix, and 
    $\mathbf{C}_{\mathbf{X}} = \mathbf{U} \boldsymbol{\Sigma} \mathbf{V}^{\mathsf T}$ is its singular value decomposition. 
    In practice, we initialize our covariance matrix with small covariance on x-y direction (5 pixel) while with large variance on z-direction (e.g., 1E+5). \\
    The major implication of this covariance analysis is a theoretical characterization of self-supervised learning. The uncertainty is high if there is small body motion as the z directional uncertainty is high. The uncertainty can be only reduced as the body move significantly where the body parts must be  seen from different viewing angles. Since our self-supervised learning leverages the expectation of the 3D point $\mathds{E}[\mathbf{X}_i]$ transformed by different time instances, the uncertainty provides a theoretical performance bound (lower bound) of self-supervised learning. This further indicates the effectiveness of self-supervised learning can be \textit{predicted} without training as the uncertainty analysis does not requires training. For instance, a sequence of a static pose would be nearly infinite along the z direction and therefore, it can be predicted that there will be no improvement with self-supervised learning.    
    \\
    \begin{table*}[th!]
  \small
  \centering
  \resizebox{2.05\columnwidth}{!}{
  \begin{tabular}{l@{\hskip 0.1in}l@{\hskip 0.1in}l@{\hskip 0.1in}l@{\hskip 0.1in}l@{\hskip 0.1in}l@{\hskip 0.1in}l@{\hskip 0.1in}l@{\hskip 0.1in}l@{\hskip 0.1in}l@{\hskip 0.1in}l@{\hskip 0.1in}l@{\hskip 0.1in}l@{\hskip 0.1in}l@{\hskip 0.1in}l@{\hskip 0.1in}l@{\hskip 0.1in}l@{\hskip 0.1in}}
  
    \toprule 
    \multicolumn{1}{c}{ }   &  \multicolumn{4}{c}{Tang et al. dataset \cite{Deephuman:2019}} &  \multicolumn{4}{c}{RenderPeople dataset \cite{RP:2020}} &       \multicolumn{4}{c}{Vlasic et al. dataset \cite{adobe_data:2008}} &       \multicolumn{4}{c}{THuman2.0 dataset \cite{tao2021function4d}} \\
     \cmidrule(r){1-1}        \cmidrule(r){2-5}                           \cmidrule(r){6-9}                               \cmidrule(r){10-13} \cmidrule(r){14-17}
    Method        & D. error           & 3cm & 4cm & 5cm   & D. error     & 3cm & 4cm & 5cm  & D. error    & 14 & 18 & 22  & D. error    & 3cm & 4cm & 5cm \\

    \cmidrule(r){1-1}\cmidrule(r){2-2} \cmidrule(r){3-3} \cmidrule(r){4-4} \cmidrule(r){5-5} \cmidrule(r){6-6} \cmidrule(r){7-7}\cmidrule(r){8-8}\cmidrule(r){9-9}\cmidrule(r){10-10}\cmidrule(r){11-11}\cmidrule(r){12-12}\cmidrule(r){13-13}\cmidrule(r){14-14}\cmidrule(r){15-15}\cmidrule(r){16-16}\cmidrule(r){17-17}
  Li et al. \cite{mannequin:2019}       &  6.1$\pm$3.2 & 4\%   & 23\%   & 48\% & 6.4$\pm$4.1   & 8\% & 28\%   & 46\% & 37.6$\pm$13.7   & 1\% & 4\%   & 8\% & 8.2$\pm$4.4   & 3\% & 11\%   & 22\% \\
 
    Tang et al. \cite{Deephuman:2019}    & \textbf{\color{blue}4.9$\pm$7.1} & \textbf{\color{red}41\%}   & \textbf{\color{red}65\%}   & \textbf{\color{red}80\%} & 6.9$\pm$2.8   & 2\% & 11\%   & 28\% & 27.1$\pm$7.9   & 1\% & 9\%   & 27\%   & 9.0$\pm$4.3   & 1\% & 6\%   & 15\% \\
    
    PIFu  \cite{pifu:2019}           & 6.3$\pm$3.4 & 5\%   & 22\%   & 46\% & 5.3$\pm$2.6   & 17\% & 35\%   & 54\% & 30.3$\pm$6.6   & 0\% & 1\%   & 8\%   & 7.3$\pm$3.3  & 3\% & 11\%   & 26\% \\
    
    PIFuHD \cite{pifuhd:2020}         & 5.5$\pm$3.0 & 11\%   & 37\%   & 57\% & 5.6$\pm$2.4   & 10\% & 29\%   & 48\% & 27.3$\pm$6.6   & 1\% & 7\%   & 23\%   & 7.8$\pm$3.9   & 3\% & 11\%   & 24\% \\
    
    PaMIR  \cite{zheng2020pamir}          & 5.4$\pm$3.0 & 12\%   & 37\%   & 58\% & 5.6$\pm$2.0   & 5\% & 21\%   & 42\% & 22.1$\pm$6.3   & 9\% & 27\%   & 53\%   & 7.2$\pm$3.2   & 3\% & 13\%   & 27\% \\
    \cmidrule(r){1-1}\cmidrule(r){2-2} \cmidrule(r){3-3} \cmidrule(r){4-4} \cmidrule(r){5-5} \cmidrule(r){6-6} \cmidrule(r){7-7}\cmidrule(r){8-8}\cmidrule(r){9-9}\cmidrule(r){10-10}\cmidrule(r){11-11}\cmidrule(r){12-12}\cmidrule(r){13-13}\cmidrule(r){14-14}\cmidrule(r){15-15}\cmidrule(r){16-16}\cmidrule(r){17-17}
  Ours \cite{Jafarian_2021_CVPR_TikTok} (affine)             & \textbf{\color{red}4.8$\pm$2.9} & \textbf{\color{blue}26\%}   & \textbf{\color{blue}52\%}   & \textbf{\color{blue}66\%} & \textbf{\color{blue}3.2$\pm$1.1}   & \textbf{\color{blue}46\%} & \textbf{\color{blue}80\%}   & \textbf{\color{blue}95\%} & \textbf{\color{blue}16.8$\pm$5.2}   & 32\% & \textbf{\color{blue}63\% }  & \textbf{\color{blue}84\%}   & \textbf{\color{red}5.1$\pm$2.5}   & \textbf{\color{red}18\%} & \textbf{\color{blue}38\%}   & \textbf{\color{red}57\%} \\
  Ours \cite{Jafarian_2021_CVPR_TikTok}  (rigid)            & 5.0$\pm$3.0 & 25\%   & 49\%   & 65\% & \textbf{\color{blue}3.2$\pm$1.1}  & \textbf{\color{blue}46\%} & \textbf{\color{blue}80\%}   & \textbf{\color{blue}95\%} & 16.9$\pm$5.1   & \textbf{\color{blue}33\%} & 61\%   & \textbf{\color{blue}84\%}   & \textbf{\color{red}5.1$\pm$2.5}   & \textbf{\color{blue}17\%} & 37\%   & \textbf{\color{red}57\%} \\
  Ours \cite{Jafarian_2021_CVPR_TikTok} + $\mathcal{L}_p$    & 5.1$\pm$2.9 & 20\%   & 47\%   & 64\% & \textbf{\color{red}2.9$\pm$1.0}   & \textbf{\color{red}56\%} & \textbf{\color{red}88\%}   & \textbf{\color{red}97\%} & \textbf{\color{red}15.9$\pm$5.4 }  & \textbf{\color{red}42\%} & \textbf{\color{red}69\%}   & \textbf{\color{red}87\%}    & \textbf{\color{blue}5.2$\pm$2.6}   & \textbf{\color{blue}17\%} & \textbf{\color{red}39\%}   & \textbf{\color{red}57\%} \\
   
    \bottomrule
  \end{tabular}}
  \vspace{-2mm}
  \caption{Quantitative results on the depth prediction. We report the depth error and the percentage of test samples having an error less than three error tolerances (3cm, 4cm, and 5cm) except for Vlasic et al. \cite{adobe_data:2008} . All the errors are reported in centimeter (cm), except for Vlasic et al. dataset \cite{adobe_data:2008} for which the conversion to metric scale is not known and the reported numbers are in their scale. The best and the second best methods are marked as \textbf{\color{red}red bold} and \textbf{\color{blue}blue bold}, respectively.
  }\label{table:Dquantitative}
\end{table*}

\begin{table*}[th!]
  \small
  \centering
  \resizebox{2.05\columnwidth}{!}{
  \begin{tabular}{l@{\hskip 0.1in}l@{\hskip 0.1in}l@{\hskip 0.1in}l@{\hskip 0.1in}l@{\hskip 0.1in}l@{\hskip 0.1in}l@{\hskip 0.1in}l@{\hskip 0.1in}l@{\hskip 0.1in}l@{\hskip 0.1in}l@{\hskip 0.1in}l@{\hskip 0.1in}l@{\hskip 0.1in}l@{\hskip 0.1in}l@{\hskip 0.1in}l@{\hskip 0.1in}l@{\hskip 0.1in}}
  
    \toprule 
    \multicolumn{1}{c}{ }   &  \multicolumn{4}{c}{Tang et al. dataset \cite{Deephuman:2019}} &  \multicolumn{4}{c}{RenderPeople dataset \cite{RP:2020}} &       \multicolumn{4}{c}{Vlasic et al. dataset \cite{adobe_data:2008}} &       \multicolumn{4}{c}{THuman2.0 dataset \cite{tao2021function4d}} \\
     \cmidrule(r){1-1}        \cmidrule(r){2-5}                           \cmidrule(r){6-9}                               \cmidrule(r){10-13} \cmidrule(r){14-17}
    Method        & N. error           & 25\degree  & 30\degree  & 35\degree    & N. error      & 25\degree  & 30\degree  & 35\degree  & N. error     & 25\degree  & 30\degree  & 35\degree  & N. error     & 25\degree  & 30\degree  & 35\degree \\

    \cmidrule(r){1-1}\cmidrule(r){2-2} \cmidrule(r){3-3} \cmidrule(r){4-4} \cmidrule(r){5-5} \cmidrule(r){6-6} \cmidrule(r){7-7}\cmidrule(r){8-8}\cmidrule(r){9-9}\cmidrule(r){10-10}\cmidrule(r){11-11}\cmidrule(r){12-12}\cmidrule(r){13-13}\cmidrule(r){14-14}\cmidrule(r){15-15}\cmidrule(r){16-16}\cmidrule(r){17-17}
  Li et al. \cite{mannequin:2019}       &  33$\pm$4 & 0\%   & 19\%   & 72\% & 28$\pm$7   & 40\% & 66\%   & 84\% & 43$\pm$8   & 2\% & 7\%   & 20\% & 32$\pm$7  & 17\% & \textbf{\color{blue}42\%}   & 67\% \\
 
    Tang et al. \cite{Deephuman:2019}    &  31$\pm$7 & \textbf{\color{red}16\%}   & 54\%   & 78\% & 35$\pm$5  & 2\% & 19\%   & 53\% & 40$\pm$7   & 0\% & 4\%   & 22\% & 39$\pm$7  & 1\% & 9\%   & 28\% \\
    
    PIFu  \cite{pifu:2019}               &  33$\pm$5 & 1\%   & 35\%   & 68\% & 25$\pm$5  & \textbf{\color{blue}51\%} & \textbf{\color{blue}79\%}   & \textbf{\color{blue}95\%} & 38$\pm$7   & 1\% & 10\%   & 34\%   & \textbf{\color{blue}32$\pm$6}  & 12\% & 40\%   & 69\% \\
    
    PIFuHD \cite{pifuhd:2020}            &  34$\pm$5 & 0\%   & 21\%   & 58\% & 27$\pm$6  & 34\% & 65\%   & 86\% & 47$\pm$6   & 0\% & 0\%   & 2\% & 35$\pm$8  & 8\% & 25\%   & 49\% \\
    
    PaMIR  \cite{zheng2020pamir}       &  34$\pm$5 & 0\%   & 15\%   & 58\% & 29$\pm$4   & 18\% & 60\%   & 87\% & 33$\pm$5   & 4\% & 29\%   & 62\% & \textbf{\color{blue}32$\pm$6} & 9\% & 40\%   & 71\% \\
    \cmidrule(r){1-1}\cmidrule(r){2-2} \cmidrule(r){3-3} \cmidrule(r){4-4} \cmidrule(r){5-5} \cmidrule(r){6-6} \cmidrule(r){7-7}\cmidrule(r){8-8}\cmidrule(r){9-9}\cmidrule(r){10-10}\cmidrule(r){11-11}\cmidrule(r){12-12}\cmidrule(r){13-13}\cmidrule(r){14-14}\cmidrule(r){15-15}\cmidrule(r){16-16}\cmidrule(r){17-17}
  Ours \cite{Jafarian_2021_CVPR_TikTok} (affine)            & \textbf{\color{red}29$\pm$4} & 8\%   & \textbf{\color{red}67\%}   & \textbf{\color{red}89\%} & \textbf{\color{blue}16$\pm$2}   & \textbf{\color{red}100\%} & \textbf{\color{red}100\%}   & \textbf{\color{red}100\%} &  \textbf{\color{red}24$\pm$4}   &  \textbf{\color{red}59\%} &  \textbf{\color{red}87\%}   &  \textbf{\color{red}97\%} &  \textbf{\color{red}22$\pm$5}  &  \textbf{\color{red}75\%} &  \textbf{\color{red}92\%}  & \textbf{\color{blue}97\%} \\
  Ours \cite{Jafarian_2021_CVPR_TikTok}  (rigid)            &  \textbf{\color{red}29$\pm$4} & \textbf{\color{blue}9\%}   & \textbf{\color{blue}65\%}   & \textbf{\color{blue}86\%} & \textbf{\color{blue}16$\pm$2}   & \textbf{\color{red}100\%} & \textbf{\color{red}100\%}   & \textbf{\color{red}100\%} & \textbf{\color{blue}25$\pm$4}   & \textbf{\color{blue}52\%} & \textbf{\color{blue}82\%}   & \textbf{\color{blue}96\%} &  \textbf{\color{red}22$\pm$5}  & \textbf{\color{blue}74\%} &  \textbf{\color{red}92\% }  &  \textbf{\color{red}98\% }\\
  Ours \cite{Jafarian_2021_CVPR_TikTok} + $\mathcal{L}_p$    &  \textbf{\color{blue}30$\pm$4} & 5\%   & 60\%   & 84\% & \textbf{\color{red}15$\pm$2}   & \textbf{\color{red}100\%} & \textbf{\color{red}100\%}   & \textbf{\color{red}100\%} & 25$\pm$5   & 49\% & 80\%   & 95\% &  \textbf{\color{red}22$\pm$5}  & \textbf{\color{blue}74\%} &  \textbf{\color{red}92\%}   &  \textbf{\color{red}98\%} \\
   
    \bottomrule
  \end{tabular}}
  \vspace{-2mm}
  \caption{Quantitative results on surface normal estimated from the depth prediction. We report the normal error and the percentage of test samples having an error less than three error tolerances (25\degree, 30\degree, and 35\degree). All the errors are reported in degree (\degree). The best and the second best methods are marked as \textbf{\color{red}red bold} and \textbf{\color{blue}blue bold}, respectively.
  }\label{table:Nquantitative}
\end{table*}

\begin{table*}[th!]
  \small
  \centering
  \resizebox{2.05\columnwidth}{!}{
  \begin{tabular}{l@{\hskip 0.1in}l@{\hskip 0.1in}l@{\hskip 0.1in}l@{\hskip 0.1in}l@{\hskip 0.1in}l@{\hskip 0.1in}l@{\hskip 0.1in}l@{\hskip 0.1in}l@{\hskip 0.1in}l@{\hskip 0.1in}l@{\hskip 0.1in}l@{\hskip 0.1in}l@{\hskip 0.1in}l@{\hskip 0.1in}l@{\hskip 0.1in}l@{\hskip 0.1in}l@{\hskip 0.1in}}
  
    \toprule 
    \multicolumn{1}{c}{ }   &  \multicolumn{4}{c}{Tang et al. dataset \cite{Deephuman:2019}} &  \multicolumn{4}{c}{RenderPeople dataset \cite{RP:2020}} &       \multicolumn{4}{c}{Vlasic et al. dataset \cite{adobe_data:2008}} &       \multicolumn{4}{c}{THuman2.0 dataset \cite{tao2021function4d}} \\
     \cmidrule(r){1-1}        \cmidrule(r){2-5}                           \cmidrule(r){6-9}                               \cmidrule(r){10-13} \cmidrule(r){14-17}
    Method        & R. error           & 3cm & 4cm & 5cm   & R. error      & 3cm & 4cm & 5cm  & R. error     & 14 & 18 & 22  & R. error     & 3cm & 4cm & 5cm \\

    \cmidrule(r){1-1}\cmidrule(r){2-2} \cmidrule(r){3-3} \cmidrule(r){4-4} \cmidrule(r){5-5} \cmidrule(r){6-6} \cmidrule(r){7-7}\cmidrule(r){8-8}\cmidrule(r){9-9}\cmidrule(r){10-10}\cmidrule(r){11-11}\cmidrule(r){12-12}\cmidrule(r){13-13}\cmidrule(r){14-14}\cmidrule(r){15-15}\cmidrule(r){16-16}\cmidrule(r){17-17}
  Li et al. \cite{mannequin:2019}       &  5.4$\pm$2.8 & 5\%   & 35\%   & 59\% & 5.1$\pm$3.0   & 16\% & 42\%   & 65\% & 27.1$\pm$10.5   & 4\% & 15\%   & 34\% & 6.9$\pm$3.6   & 7\% & 18\%   & 34\% \\
 
    Tang et al. \cite{Deephuman:2019}    & 4.6$\pm$6.8 & \textbf{\color{red}47\%}   &  \textbf{\color{red}71\%}   & \textbf{\color{red}83\%} & 5.8$\pm$2.3   & 6\% & 19\%   & 42\% & 22.9$\pm$7.2   & 6\% & 25\%   & 54\%   & 7.6$\pm$3.6   & 2\% & 10\%   & 24\% \\
    
    PIFu  \cite{pifu:2019}           & 5.6$\pm$3.3 & 11\%   & 41\%   & 58\% & 4.2$\pm$2.0   & 29\% & 53\%   & 72\% & 22.3$\pm$6.9   & 8\% & 30\%   & 53\%   & 6.2$\pm$2.9   & 7\% & 23\%   & 41\% \\
    
    PIFuHD \cite{pifuhd:2020}         & 4.9$\pm$2.8 & 21\%   & 49\%   & 65\% & 4.3$\pm$1.7   & 24\% & 51\%   & 73\% & 21.5$\pm$5.8   & 7\% & 29\%   & 59\%   & 6.5$\pm$3.3   & 6\% & 22\%   & 38\% \\
    
    PaMIR  \cite{zheng2020pamir}          & 4.9$\pm$2.9 & 21\%   & 48\%   & 67\% & 4.7$\pm$1.5   & 10\% & 33\%   & 62\% & 17.0$\pm$5.5   & 32\% & 64\%   & 83\%   &6.0$\pm$2.8   & 7\% & 23\%   & 41\% \\
    \cmidrule(r){1-1}\cmidrule(r){2-2} \cmidrule(r){3-3} \cmidrule(r){4-4} \cmidrule(r){5-5} \cmidrule(r){6-6} \cmidrule(r){7-7}\cmidrule(r){8-8}\cmidrule(r){9-9}\cmidrule(r){10-10}\cmidrule(r){11-11}\cmidrule(r){12-12}\cmidrule(r){13-13}\cmidrule(r){14-14}\cmidrule(r){15-15}\cmidrule(r){16-16}\cmidrule(r){17-17}
  Ours \cite{Jafarian_2021_CVPR_TikTok} (affine)             & \textbf{\color{red}4.4$\pm$2.6} & 31\%   & \textbf{\color{blue}59\%}   & 72\% & \textbf{\color{blue}2.8$\pm$0.9}   & 61\% & 89\%   & 96\% & \textbf{\color{red}12.8$\pm$4.4}   & \textbf{\color{blue}64\%} & \textbf{\color{red}88\%}   & \textbf{\color{red}96\%}   & \textbf{\color{red}4.4$\pm$2.2}   & \textbf{\color{blue}29\%} & \textbf{\color{red}53\%}   & \textbf{\color{red}71\%} \\
  Ours \cite{Jafarian_2021_CVPR_TikTok}  (rigid)            & \textbf{\color{blue}4.5$\pm$2.7} & \textbf{\color{blue}33\%}   & 58\%   & 72\% & \textbf{\color{blue}2.8$\pm$0.9}   & \textbf{\color{blue}63\%} & \textbf{\color{blue}90\%}   & \textbf{\color{blue}97\%} & \textbf{\color{blue}13.1$\pm$4.6}   & 62\% & \textbf{\color{blue}86\%}   & \textbf{\color{blue}95\%}   &  \textbf{\color{red}4.4$\pm$2.2}   & 28\% & \textbf{\color{blue}52\% }  & \textbf{\color{blue}70\%} \\
  Ours \cite{Jafarian_2021_CVPR_TikTok} + $\mathcal{L}_p$      & \textbf{\color{red}4.4$\pm$2.6} & 32\%   & \textbf{\color{blue}59\%}   & \textbf{\color{blue}73\%} & \textbf{\color{red}2.5$\pm$0.8}   & \textbf{\color{red}71\%} & \textbf{\color{red}93\%}   & \textbf{\color{red}99\%} & \textbf{\color{red}12.8$\pm$4.7}   & \textbf{\color{red}66\%} & \textbf{\color{blue}86\% }  & \textbf{\color{blue}95\%}    & \textbf{\color{blue}4.4$\pm$2.3}   & \textbf{\color{red}30\%} & \textbf{\color{red}53\%}   & \textbf{\color{blue}70\%} \\
   
    \bottomrule
  \end{tabular}}
  \vspace{-2mm}
  \caption{Quantitative results on surface reconstruction. We report the reconstruction error and the percentage of test samples having an error less than three error tolerances (3cm, 4cm, and 5cm) except for Vlasic et al. \cite{adobe_data:2008}. All the errors are reported in centimeter (cm), except for Vlasic et al. dataset \cite{adobe_data:2008} for which the conversion to metric scale is not known and the reported numbers are in their scale. The best and the second best methods are marked as \textbf{\color{red}red bold} and \textbf{\color{blue}blue bold}, respectively.}
  \label{table:Rquantitative}
\end{table*}

    \noindent\textbf{Experimental Validation:}  We compare the reconstruction uncertainty in relation with the depth error using short sequences of the Tang et. al. dataset \cite{Deephuman:2019}: Ex1 made of frame 1-5, Ex2 made of 1-10, Ex3 made of 1-15, Ex4 made of 1-20, Ex5 made of 1-25, and Ex6 made of 1-30. Ex1 is a short sequence with small body displacement that is expected to be high uncertainty while Ex6 is a long sequence with full body rotation that is expected to be low uncertainty. In Figure~\ref{Fig:unceVsErr}, we illustrates the depth error produced by self-supervised learning as a function of the uncertainty for torso, head, arms, and legs. As expected, the depth error after self-supervised learning increases as the uncertainty increases.


\section{Experiments}

We evaluate our method both quantitatively and qualitatively compared with the state-of-the-art methods of human depth estimation and human shape recovery on real and synthetic data.

\begin{table*}[t]
  \normalsize
  \centering
  \begin{tabular}{l@{\hskip 0.1in}l@{\hskip 0.1in}l@{\hskip 0.1in}l@{\hskip 0.1in}}
  
     \toprule                           
    Losses                    & D. error           & N. error   & R. error      \\

    \cmidrule(r){1-1}\cmidrule(r){2-2} \cmidrule(r){3-3} \cmidrule(r){4-4} 
   $\mathcal{L}_\mathbf{z}$ (trained on RenderPeople dataset \cite{RP:2020})        & {5.66$\pm$3.85}   & {34.24$\pm$5.72}  & {5.17$\pm$3.09}   \\
 
    $\mathcal{L}_\mathbf{z}+\mathcal{L}_\mathbf{s}$ (trained on RenderPeople dataset \cite{RP:2020})    & {5.11$\pm$3.20}   & {29.99$\pm$4.85}  & {4.66$\pm$2.82}    \\
    
      $\mathcal{L}_\mathbf{z}+\mathcal{L}_\mathbf{s}+\mathcal{L}_\mathbf{w}$ (trained on RenderPeople \cite{RP:2020} and TikTok dataset)            & \textbf{4.89$\pm$2.93}   & \textbf{29.36$\pm$4.40}  & \textbf{4.46$\pm$2.65}   \\
   
    \bottomrule
  \end{tabular}
  \vspace{1mm}
  \caption{Ablation study on Tang et al. dataset \cite{Deephuman:2019}. Here we report the depth error (cm), normal error (\degree) and reconstruction error (cm) (mean$\pm$std). }
  \label{table:ablation}
\end{table*}

\noindent\textbf{Training Datasets} We use two datasets for training: 340 subjects from 3D scanned model (RenderPeople)~\cite{RP:2020} with 3D ground truth and our TikTok dataset without 3D ground truth (Section~\ref{Sec:dataset}). We render the  3D scanned mesh models from approximately 100 viewpoints sampled uniformly across a camera rig (6m diameter) that encircles each subject with 16.5mm focal length. Total 34,000 and 100,000 images are used for training from RenderPeople and TikTok data, respectively. 

\noindent\textbf{Evaluation Datasets} We use four datasets to compare the performance of ours and baseline methods: Tang et al.~\cite{Deephuman:2019}, RenderPeople~\cite{RP:2020}, Vlasic et al.~\cite{adobe_data:2008}, and THuman2.0~\cite{tao2021function4d}.
1) \textit{Training dataset of Tang et al.} This dataset is made of sequences of depth and RGB image pair for 25 subjects. We randomly choose around 70 frames for each subject, totaling 1300 images. 2) \textit{RenderPeople dataset} This dataset is made of 3D scanned models with texture. We choose 6 subjects that are not part of our training data and render the images from 100 viewpoints, totaling 600 images. We use a ray tracing algorithm to compute the ground truth depth and render the human textured model. 3) \textit{Vlasic et al. dataset} This dataset consists of 10 sequences of different people viewed from 8 views. Each video includes average of 200 frames of diverse activities such as swing dancing, samba dancing, jumping, squat, and marching. The dataset provides the RGB images and the meshes along with the camera parameters. We use a ray tracing algorithm to generate the ground truth depth from the meshes. We randomly choose total of 2000 images from this dataset. This dataset is in particular challenging because the viewpoints are substantially different from the existing datasets, i.e., a subject is viewed from an oblique view. 
4) \textit{THuman2.0 dataset} This dataset consists of 526 3D scanned models with texture. We use a ray tracing algorithm to compute the ground truth depth and render the human textured model. We randomly chose 140 3D models and rendered them on 15 cameras around them, to generate totaling 2100 test images.

\begin{figure}[t]
\begin{center}
\subfigure[Impact on self-supervision]
    {\label{Fig:qual_ablation}\includegraphics[width=0.95\linewidth]{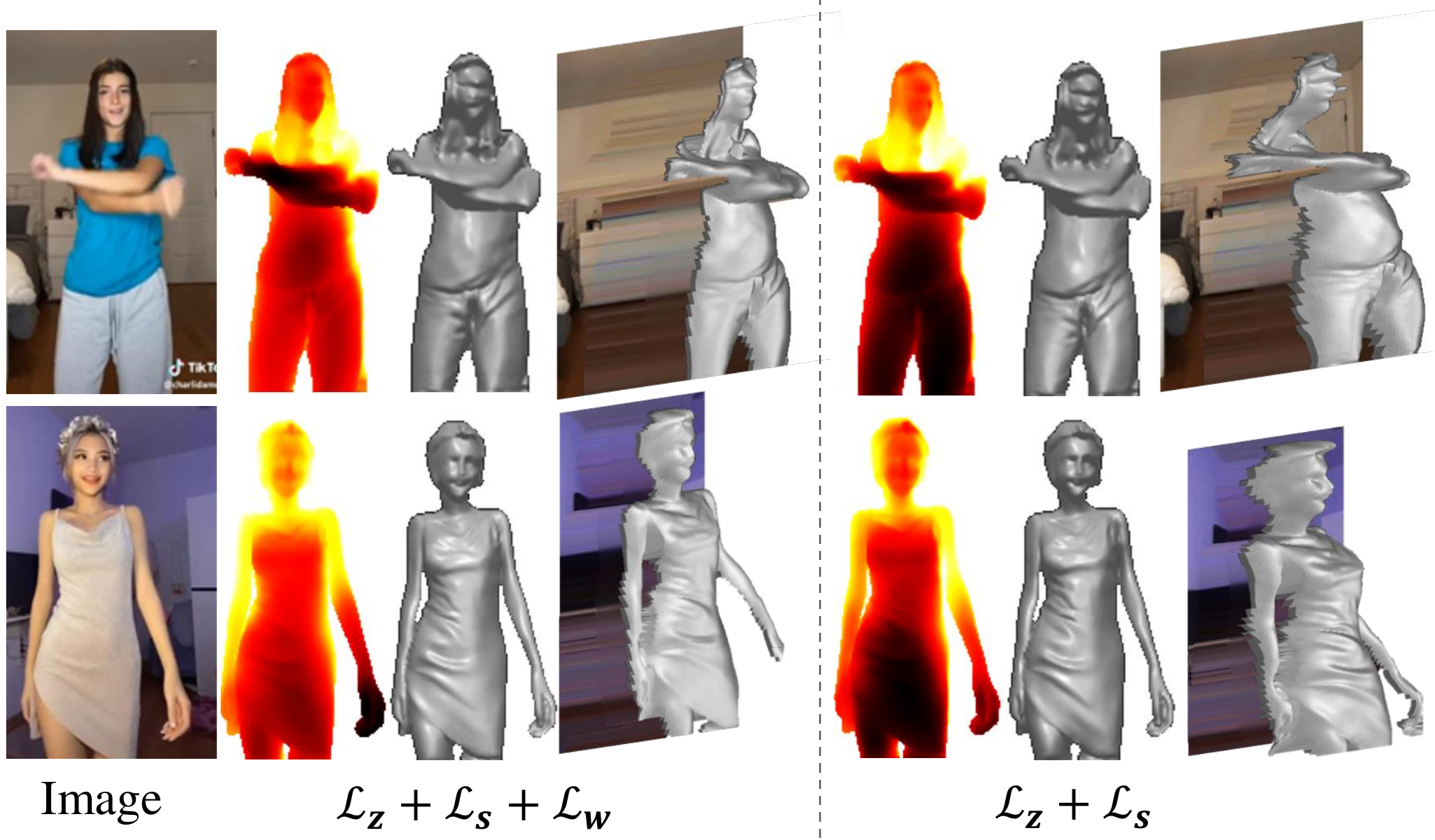}}\\
    \vspace{-3mm}
\subfigure[Depth error histogram.]
    {\label{Fig:error_histo}\includegraphics[width=0.95\linewidth]{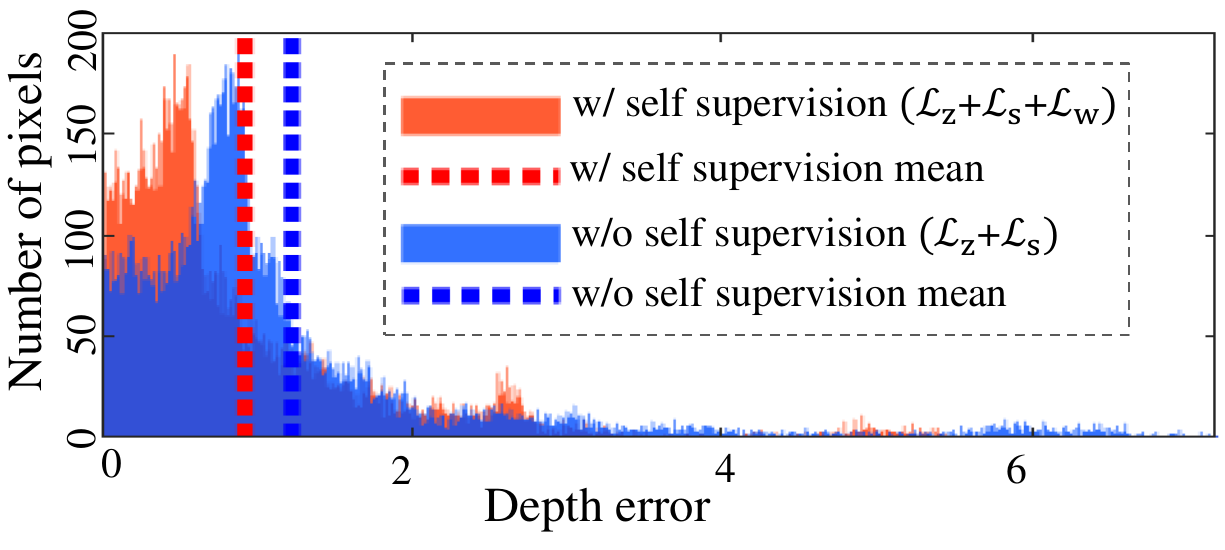}}
\end{center}
\vspace{-5mm}
   \caption{(a) Ablation study on loss functions. From left to right: the image, the full method results, and the results without self-supervision. (b) We show the depth error histogram that illustrates the long tail error distribution of the models without self-supervision.}
\label{Fig:qunt_histo_all}
\end{figure}


\noindent\textbf{Evaluation Metric} 
We evaluate the performance in two aspects: (1) accuracy of depths, surface normals, and 3D reconstruction, and (2) impact of joint training of surface normal and depth ($\mathcal{L}_{s}$) and integration of real dance videos ($\mathcal{L}_{w}$). 
We use mean squared error and mean absolute angular error as a metric for depth (Table \ref{table:Dquantitative}) and surface normal (Table \ref{table:Nquantitative}), respectively. The surface normals are computed via Equation~(\ref{Eq:pde}) and compared with the ground truth.  In addition, we measure the 3D error by reconstructing 
3D point cloud from the estimated depths. To handle unknown scale and focal length, we estimate a relative transformation between the estimated 3D point cloud and the ground truth to measure the shape error, i.e., the estimated point cloud is translated to the median of ground truth and scaled to match the minimum/maximum point cloud distance. The reconstruction error is computed using mean square error (Table \ref{table:Rquantitative}).

\subsection{Quantitative Evaluation}

 We followed the evaluation protocol of Li et al. \cite{mannequin:2019}, i.e., no retraining of the baseline models. We categorize the baseline methods into two: human depth estimation~\cite{mannequin:2019, Deephuman:2019}, and human shape recovery~\cite{pifu:2019, pifuhd:2020, zheng2020pamir}. The quantitative comparison is summarized in Table~\ref{table:Dquantitative},~\ref{table:Nquantitative},~\ref{table:Rquantitative}. We report the performance of our method for rigid transformation warping (first row), affine transformation warping (second row), and the effect of photometric consistency ($\mathcal{L}_\mathbf{p}$) (last row).


\noindent i) \textit{Human shape recovery} We compare our method with non-parametric human shape recovery designed for dressed humans (PIFu \cite{pifu:2019}, PIFuHD \cite{pifuhd:2020}, and PaMIR \cite{zheng2020pamir}) using an implicit function. Note that these methods predict not only the frontal body surface but also occluded body surface where we measure error only for the visible region. We apply a ray tracing method to identify the frontal surface where we measure the depth and surface normal. 

\noindent ii) \textit{Human depth estimation} We compare with depth estimation baselines that are tailored to dressed humans, which are most relevant to our work. Li et al. \cite{mannequin:2019} used a large community dataset called MannequinChallenge dataset to train the stacked hourglasses~\cite{newell:2006}, and Tang et al. \cite{Deephuman:2019} leveraged surface normals and depths to preserve detailed dressed human shapes. 
Note that Tang et al. \cite{Deephuman:2019} is both trained and tested on the Tang et al. dataset (no testing data are provided). This results in strong performance of Tang et al., which forms an upper bound performance on Tang et al. dataset. Nonetheless, our method without any adaptation to the dataset performs competitively and generalizes well.

    As shown in Table~\ref{table:Dquantitative},~\ref{table:Nquantitative},~\ref{table:Rquantitative}, affine transformation is more expressive, in general, compared to rigid body transformation. This is due to the ability to model nonrigid transformation to some extent where most body parts undergo nonrigid motion. Therefore, as expected results of the affine transformation generally surpass the rigid transformation. For future research, the warping solving in our framework can be further extended to perspective transformation or even nonlinear fittings for more degree of freedom and a better warping representation.\\
    As reported in Table~\ref{table:Dquantitative},~\ref{table:Nquantitative},~\ref{table:Rquantitative}, except in the RenderPeople dataset, the photometric loss ($\mathcal{L} p$) is not effective as other losses. This stems from the fact that human appearance in RenderPeople is well textured compared to other datasets, which can benefit from the photometric coherence.




\begin{figure}[t]
  \begin{center}
  \vspace{-4mm}
    \hspace{-5mm}
    \includegraphics[width=0.51\textwidth]{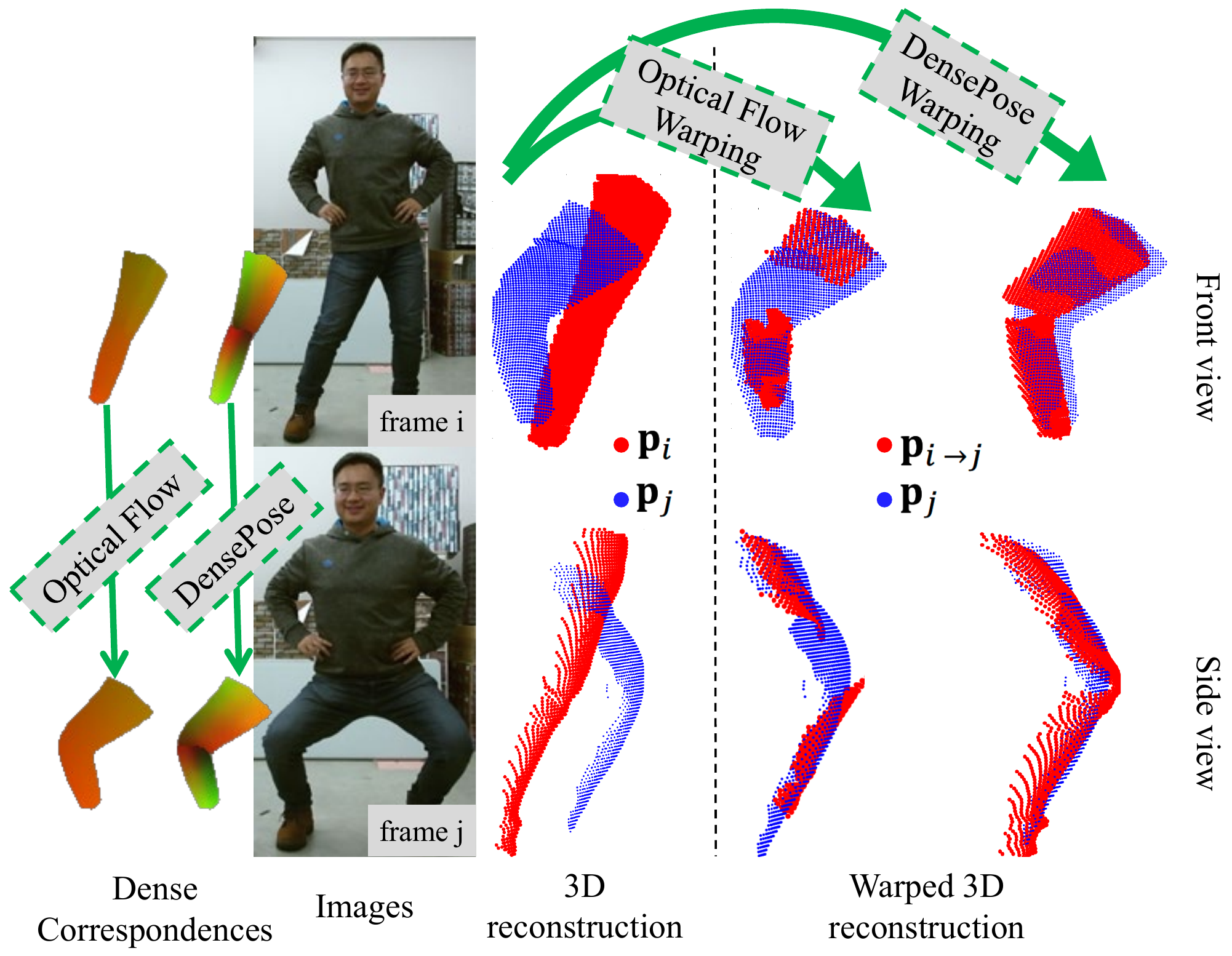}
  \end{center}
     \vspace{-4mm}
  \caption{Comparison between DensePose and optical flow correspondences and their estimated warping on two frames (3 frames apart) of a sequence. From left to right we show (1) the correspondences from frame $i$ to $j$ using optical flow, (2) the correspondences from frame $i$ to $j$ using DensePose, (3) images in frame $i$ and $j$, (4) the point cloud reconstruction from the predicted depth, in frame $i$ and $j$, from the front and side view, (5) the warped point cloud from $i$ to $j$ using optical flow correspondences, and point cloud in frame $j$, and (6) the warped point cloud from $i$ to $j$ using DensePose correspondences, and point cloud in frame $j$.}
  \label{Fig:OF_DP}
\end{figure}

\subsection{Ablation Study} 


We conduct an ablation study to analyze the impact of the losses and the usage of TikTok videos in training: $\mathcal{L}_\mathbf{z}$ , $\mathcal{L}_\mathbf{w}$ and $\mathcal{L}_\mathbf{s}$. We consider three combinations: $\mathcal{L}_\mathbf{z}$, $\mathcal{L}_\mathbf{z}$+$\mathcal{L}_\mathbf{s}$, and $\mathcal{L}_\mathbf{z}$+$\mathcal{L}_\mathbf{s}$+$\mathcal{L}_\mathbf{w}$. We use the Tang et al. dataset \cite{Deephuman:2019} without an adaptation for the evaluation. We scale the predicted depths to match to the ground truth, i.e., the predicted depths are translated to the median of ground truth and scaled to match the minimum/maximum depths.
Table~\ref{table:ablation} summarizes the comparison of the combinations.
The first two rows in Table \ref{table:ablation} is trained on only the RenderPeople dataset as 
we have the
ground truth depth and surface normal. The last row is trained on RenderPeople 
and the TikTok dataset 
together.
Note that $\mathcal{L}_w$ and $\mathcal{L}_p$ can be only applied to TikTok data to leverage motion. The RenderPeople dataset we had access to was 3D posed data which only captures the human mesh in one frame so we do not have access to any other pose of that mesh; thus, we cannot apply $\mathcal{L}_w$ and $\mathcal{L}_p$ on this dataset.
On the one hand, $\mathcal{L}_\mathbf{w}$ enforces the network to learn the geometric consistency from the videos. This loss is highly effective and allows learning from a limited amount of 3D data. On the other hand, $\mathcal{L}_\mathbf{s}$ enforces to learn to recover the details, which can further reduce the depth and surface normal errors. Our method that leverages all three losses shows the most accurate prediction in reconstructing the depths and surface normals (last row of Table ~\ref{table:ablation}).

Our self-supervision makes a positive impact on the plausibility of reconstruction. Without it, the trained model is highly
over-fitted to the scanned data, which produces unrealistic reconstruction as shown in Figure  \ref{Fig:qual_ablation}. From left to right we have the  image, the final method results and the results without self supervision.  Without the self supervision, The head is reconstructed far behind the torso mainly due to the small size of the head.
As the mean and median errors are not the best descriptive metrics to capture such qualitative plausibility, we further analyze the error by computing its distribution using error histogram shown in Figure \ref{Fig:error_histo}. 
The self-supervision results in majority of pixels remaining in the lower error regions and a smaller number of pixels in outlier regions (shorter tail error distribution).

\noindent
\textbf{Comparison with Optical Flow:} To examine the choice of DensePose as the underlying dense correspondences in our framework, we conduct a new comparison of body part warping using DensePose and optical flow. Optical flow that is designed for small pixel displacement fails to make correspondences for far distant frames (e.g., more than 10 frames) while DensePose can be applied regardless of frame distance.  
Nonetheless, Figure \ref{Fig:OF_DP} shows the 3D warping based on DensePose and optical flow for the right leg in the Tang et al. dataset (3 frames apart). 
It illustrates the warping from the DensePose produces accurate alignment of point cloud that can cover the entire leg.

\begin{figure*}[t]
  \begin{center}
    \includegraphics[width=\textwidth]{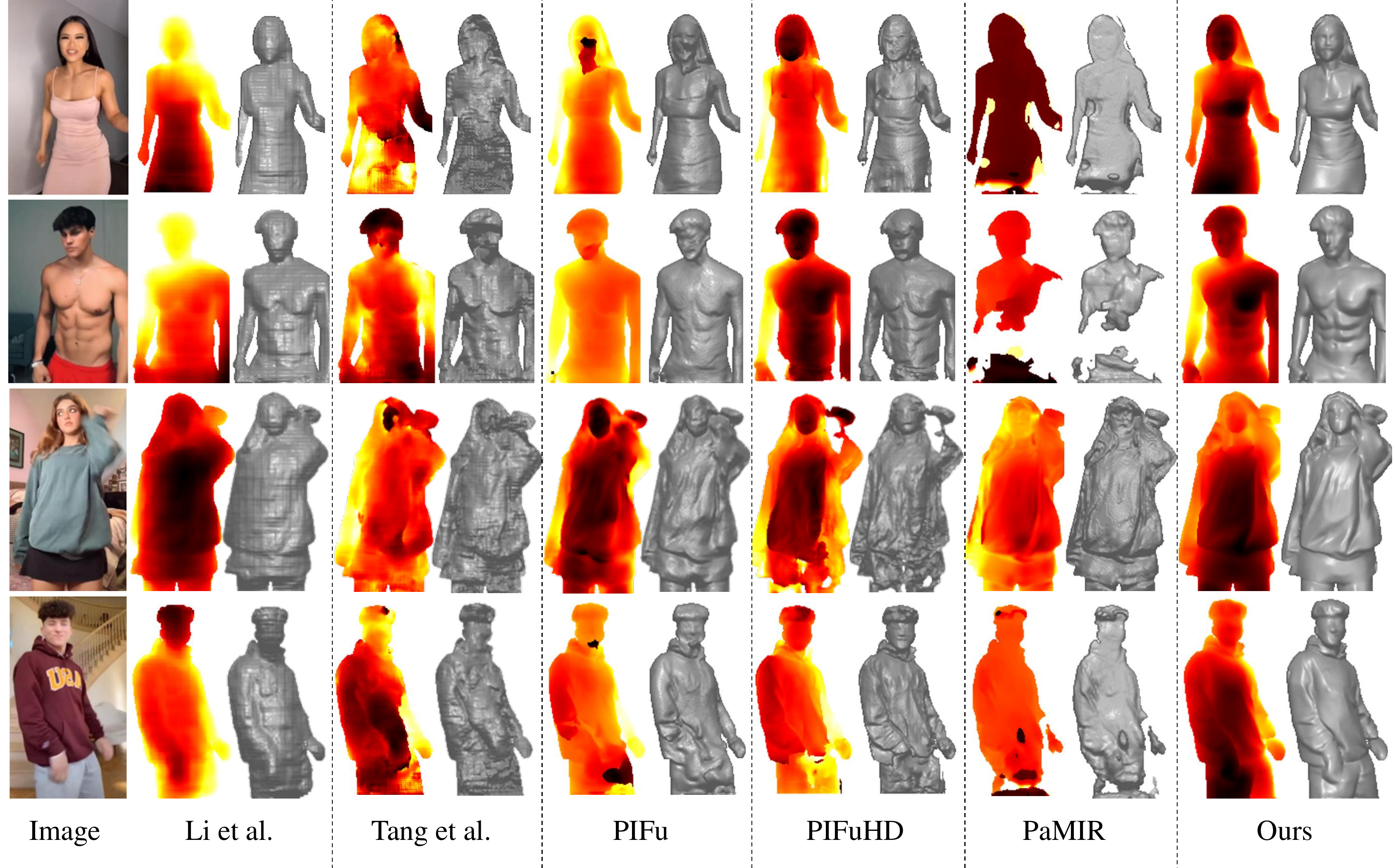}
  \end{center}
  \caption{Qualitative comparison on TikTok dataset \cite{Jafarian_2021_CVPR_TikTok}.}
  \label{Fig:qual_tk}
  \vspace{3mm}
\end{figure*}

\begin{figure*}[h!]
\begin{center}
\subfigure[Generalization to internet images]
    {\label{Fig:qual_web}\includegraphics[height=0.5\textwidth]{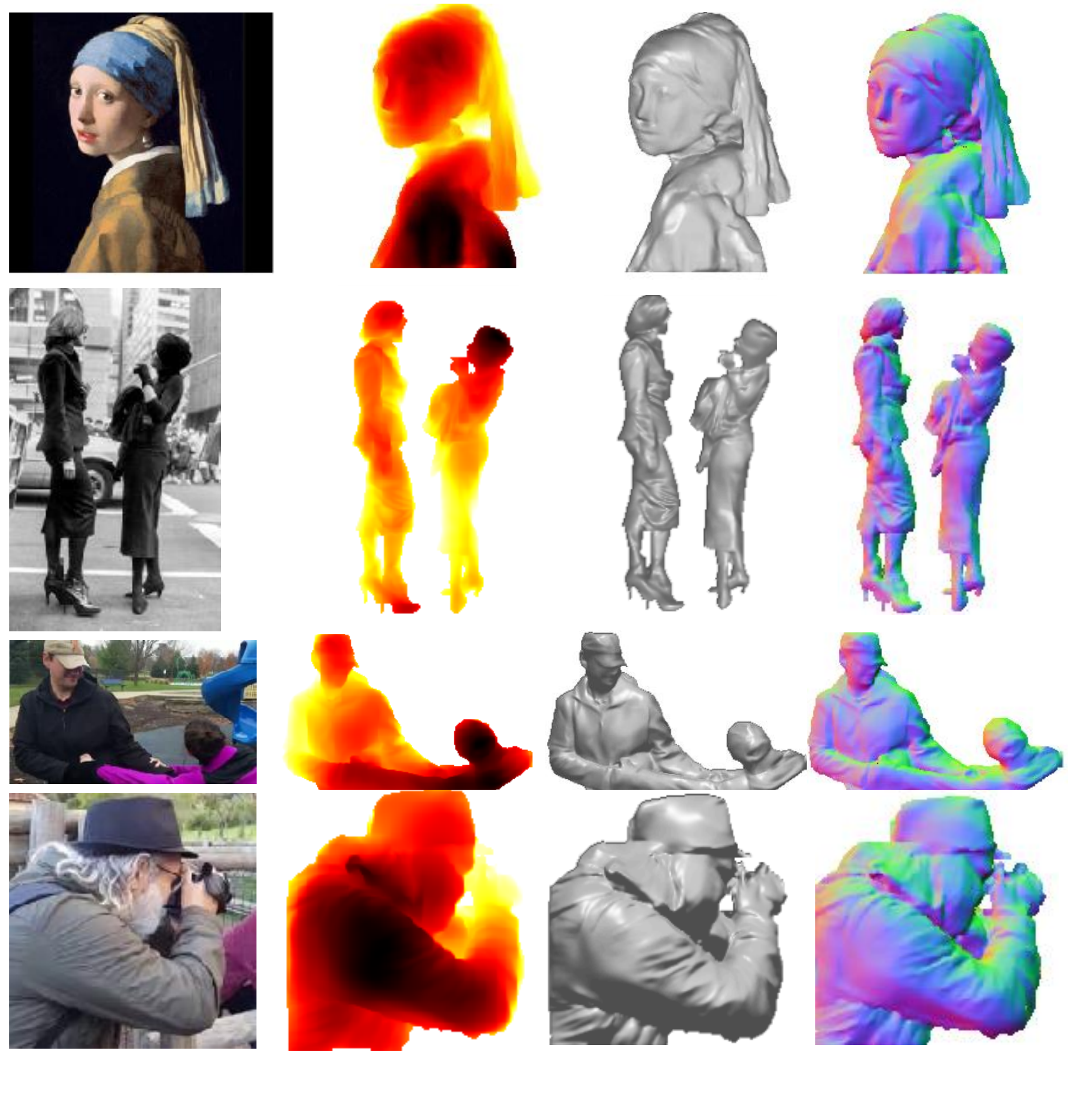}}~~~
\subfigure[Large pose transformation]
    {\label{Fig:left_leg}\includegraphics[height=0.5\textwidth]{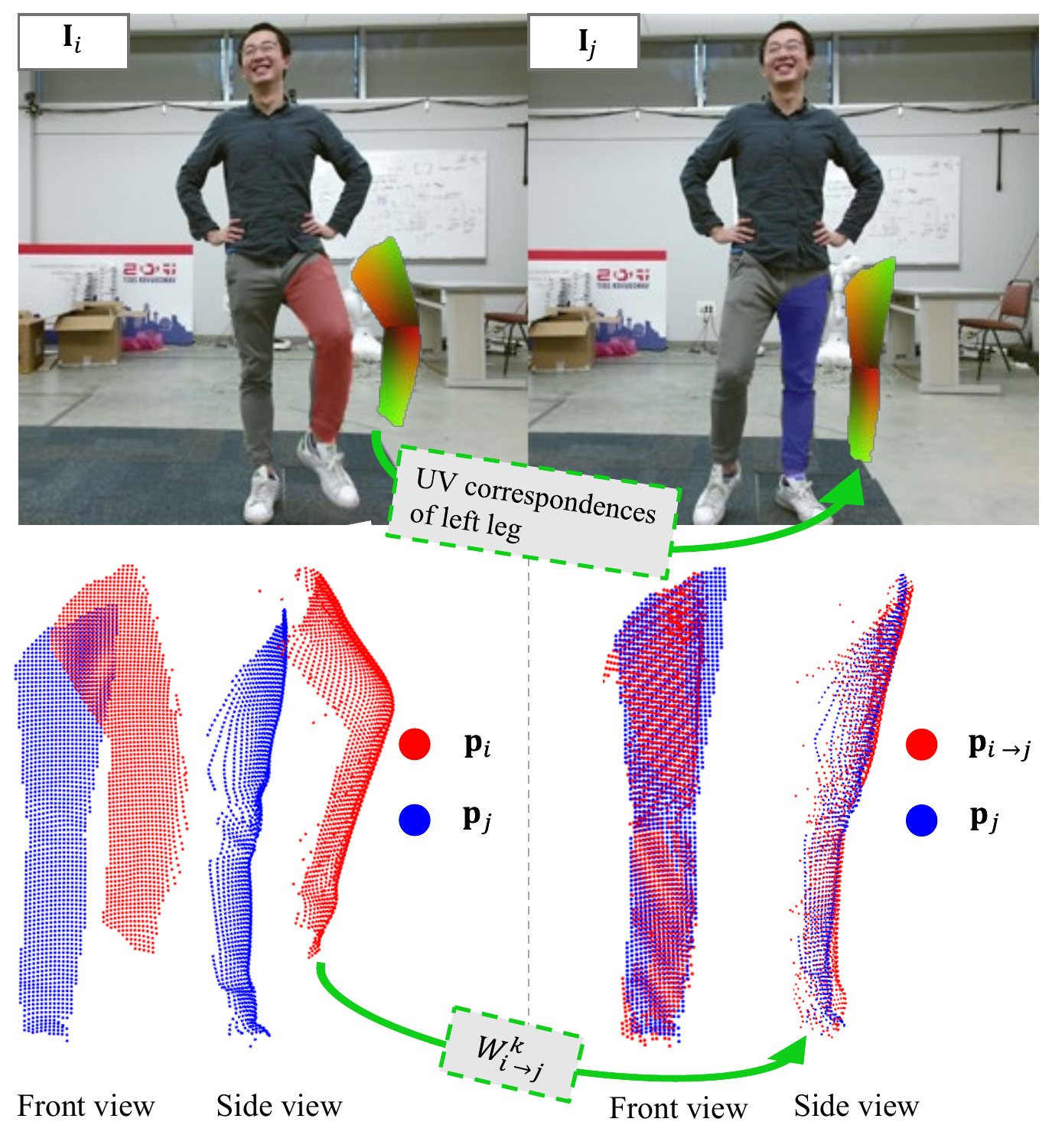}}
\end{center}
\vspace{-3mm}
   \caption{(a) Qualitative results of our method on web images. From left to right: image, predicted depth, reconstructed surface and surface normal. (b) Our method can handle significant depth change.}
\label{Fig:qual_all1}
\end{figure*}
\begin{figure*}[th!]
\begin{center}
\subfigure[Qualitative evaluation on Tang et al. dataset \cite{Deephuman:2019}.]
    {\label{Fig:qual_tg}\includegraphics[width=0.92\textwidth]{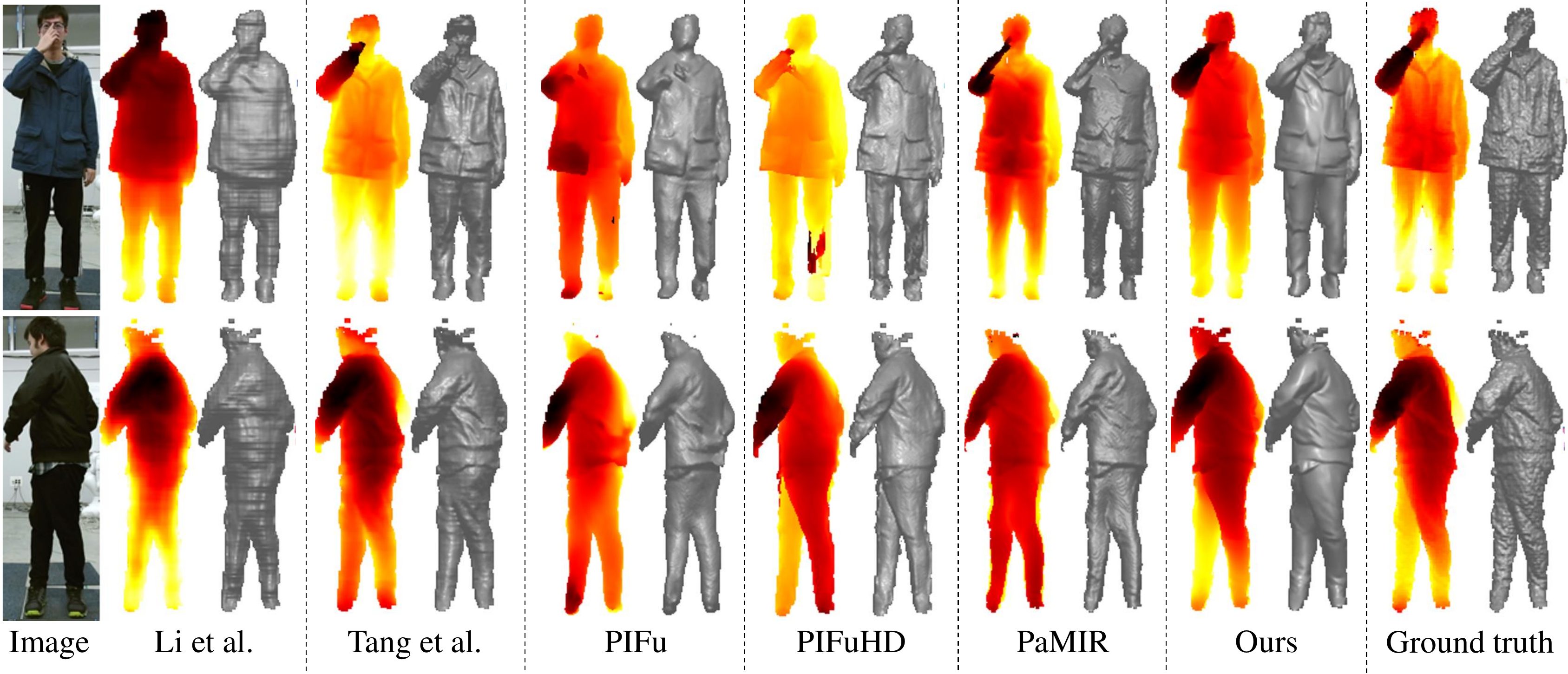}}\\
\subfigure[Qualitative evaluation on RenderPeople dataset \cite{RP:2020}.]
    {\label{Fig:qual_rp}\includegraphics[width=0.92\textwidth]{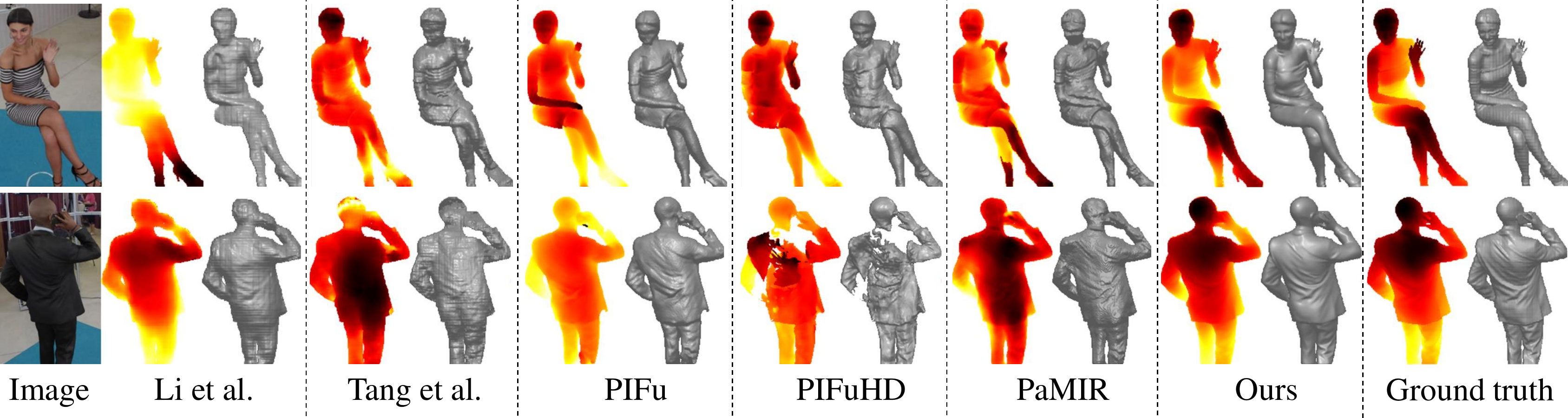}}\\
\subfigure[Qualitative evaluation on Vlasic et al. dataset \cite{adobe_data:2008}.]
    {\label{Fig:qual_dd}\includegraphics[width=0.92\textwidth]{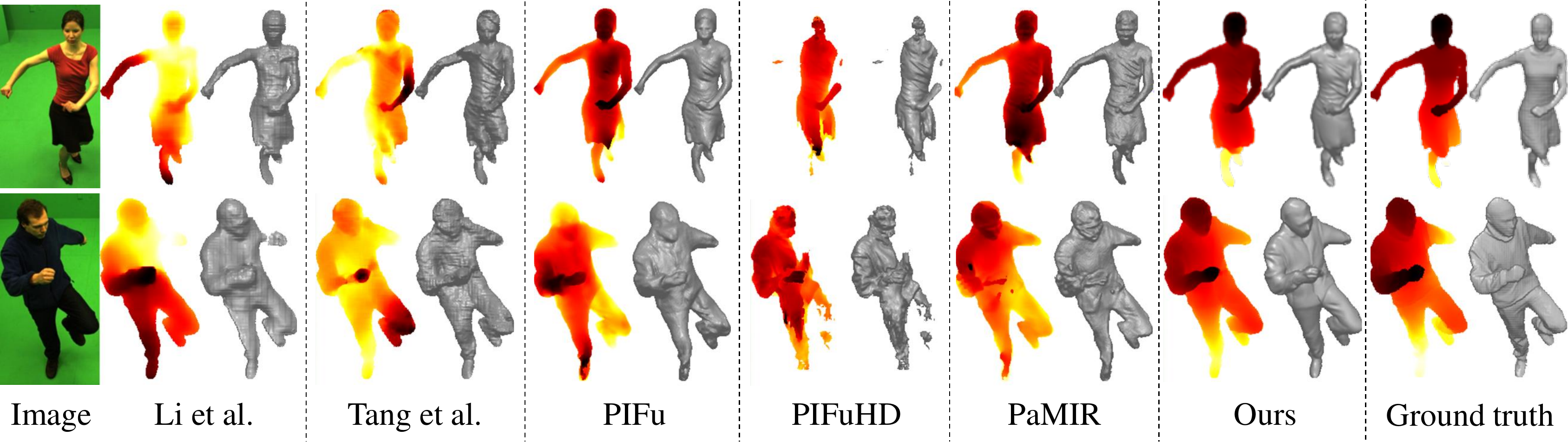}}\\
\subfigure[Qualitative evaluation on THuman2.0 dataset \cite{tao2021function4d}.]
    {\label{Fig:qual_th}\includegraphics[width=0.92\textwidth]{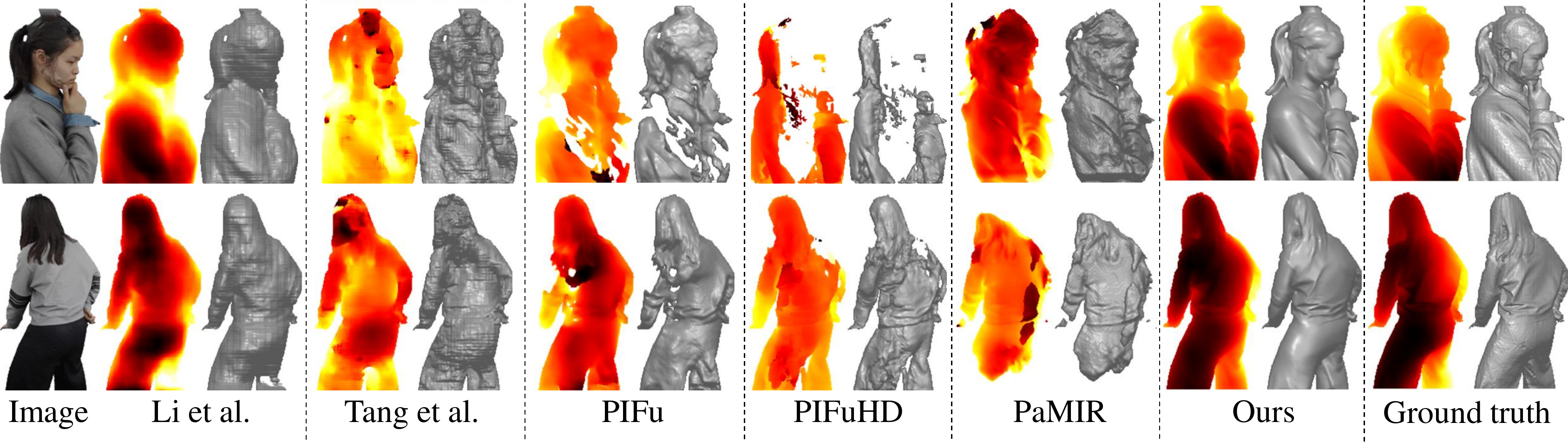}}\\
\end{center}
\vspace{-3mm}
   \caption{Qualitative results of our method and baselines on 4 different evaluation dataset.}
\label{Fig:qual_all}
\vspace{-5mm}
\end{figure*}

\subsection{Qualitative Evaluation} 

Figure \ref{Fig:qual_tk} shows the evaluation of our method compared to the baseline methods on TikTok dataset \cite{Jafarian_2021_CVPR_TikTok}. We get the most representative depth estimation compared to other methods. 

We visualize the performance of our method on a set of web images in Figure \ref{Fig:qual_web}. Our method is generalizable to gray scale images, paintings, and images with multiple people. 

We also evaluate our method compared to the baselines qualitatively on the evaluation datasets (Figure \ref{Fig:qual_all}). 
Figure \ref{Fig:qual_tg} shows the performance of our method and the baselines compared to the ground truth on Tang et al. dataset \cite{Deephuman:2019}. Note that Tang et al. \cite{Deephuman:2019} was trained on this data. Our method has the most plausible results compared to the baselines on this dataset.
Figure \ref{Fig:qual_rp} shows the performance of our method and the baselines compared against the ground truth on RenderPeople dataset \cite{RP:2020}. Our prediction is the closest to the ground truth compared to the baseline methods.
Figure \ref{Fig:qual_dd} and \ref{Fig:qual_th} also show that our method outperforms the baselines on Vlasic et al. \cite{adobe_data:2008} data and THuman2.0 data \cite{tao2021function4d} respectively.

\noindent\textbf{Handling Large Pose Variation} Our method can handle moving body parts, such as arms and legs, that induce significant depth variation across time. Specifically, the 3D translation in $\mathcal{W}_{i\rightarrow j}^k$ is designed to account for such changes in depth. Figure \ref{Fig:left_leg} shows
a large depth and pose change of the left leg between frames where the 3D points ($\mathbf{p}_{i}$) can be correctly transformed to the other frame ($\mathbf{p}_{i \rightarrow j} = \mathcal{W}_{i\rightarrow j}^k(\mathbf{p}_i)$).

\subsection{{Failure Cases}}

\begin{figure*}
  \begin{center}
    \includegraphics[width=0.92\textwidth]{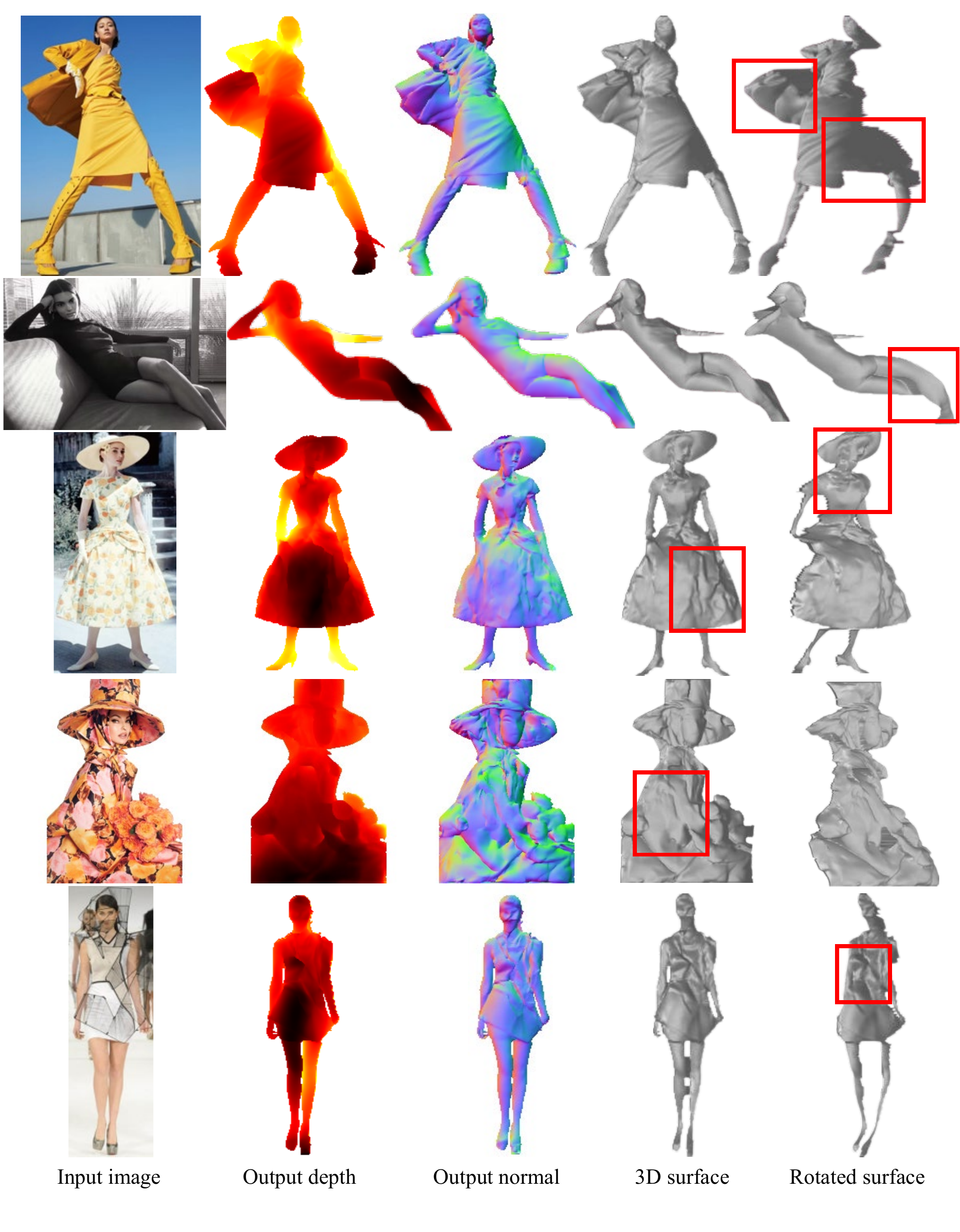}
  \end{center}
     \vspace{-6mm}
  \caption{Failure cases. Here we show the input image, predicted depth, surface normal, and the reconstructed 3D surface from the front and side view. Despite robustness in most of the images in the wild, in some extreme cases our method can fail to predict a realistic depth. These failure scenarios can be categorized as: 1)\textbf{ uncommon camera view:} first row (\textit{credit: Max Mara Clothing on Instagram}), 2) \textbf{extreme lighting and unnatural coloration:} second row (\textit{credit: Kendall Jenner on Instagram}) and third row (\textit{credit: Audrey Hepburn in FUNNY FACE 1956 Paramount film}), 3) \textbf{highly crowded texture:} third row and fourth row (\textit{credit: Linda Evangelista's British Vogue Interview: The Rebirth Of The Indomitable Super}), and 4) \textbf{occluding accessories:}  fifth row (\textit{credit: Collection by Richard Sun of UCA Rochester. Graduate Fashion Week 2012 at London's Earl's Court.}) 
}
  \label{Fig:failcases}
\end{figure*}

As our method takes advantage of in-the-wild internet data, it is robust to the majority of different viewpoints and appearances. However, in some extreme and rare cases, our method fails to predict a realistic human depth. The failure scenarios (shown in Figure \ref{Fig:failcases}) are caused by one of these factors: 1)\textbf{ uncommon camera view:} first row (The left leg is predicted way further from the body and the bottom of the coat is predicted closer to the camera because of the unusual point of view), 2) \textbf{extreme lighting and unnatural coloration:} second row (the left lower part of the leg is predicted curved because of the lighting of the room) and third row (the shadow of the hat over the body made the depth prediction of the head and neck unrealistic), 3) \textbf{highly crowded texture:} third row and fourth row (the texture on both of the dresses can be miss-identified as shadows in normal estimator and lead to uneven reconstruction), and 4) \textbf{occluding accessories:}  fifth row (the net accessories around the dress and head can cause confusion for the normal estimator in identifying the surface and lead to uneven reconstruction).

\section{Conclusion}
This paper presents a new method to utilize large data of video data shared in social media to predict the depths of dressed humans. Our formulation allows self-supervision of depth prediction by leveraging local transformations to enforce geometric consistency across different poses. In addition, we jointly learn the surface normal and depth to generate high fidelity depth reconstruction. A new dataset called TikTok dataset is collected, consisting of 340 sequences of dance videos shared in a social media mobile platform, TikTok, totaling more than 100K images. Our method produces strong qualitative and quantitative prediction on real world imagery compared to the state-of-the-art human depth estimation and human shape recovery. \\
\vspace{-2mm}

\noindent\textbf{Acknowledgement}
This work was supported by a NSF NRI 2022894 and NSF  CAREER 1846031.



{\small
\bibliographystyle{ieee}
\bibliography{egbib}
}
\newpage

\begin{IEEEbiography}[{\includegraphics[width=1in,height=1.25in,clip,keepaspectratio]{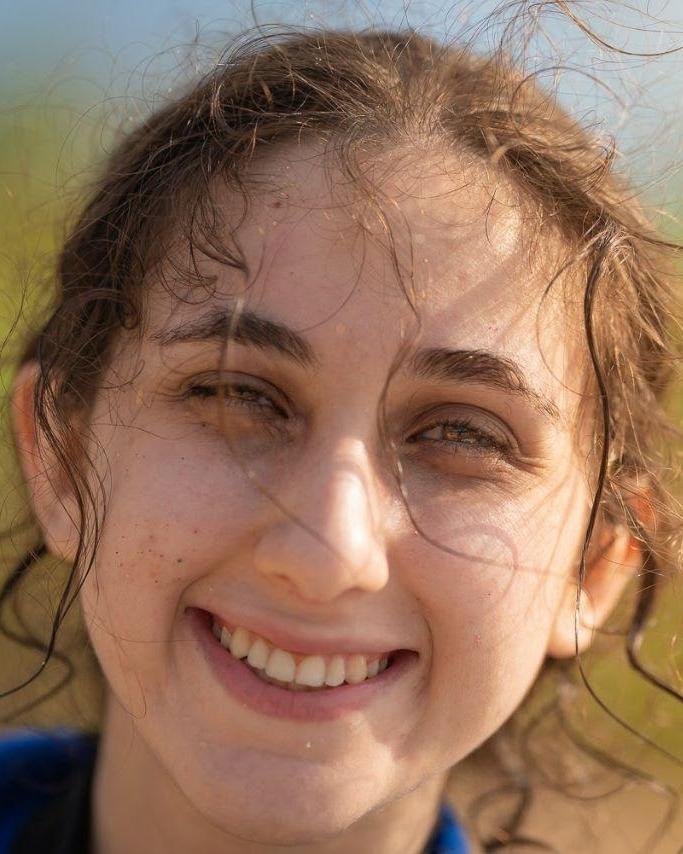}}]{Yasamin Jafarian}
Yasamin Jafarian is a Ph.D. candidate in the Department of Computer Science and Engineering at the University of Minnesota. She is interested in understanding the high fidelity 3D human geometry from single view images using self-supervised learning. She received the CVPR 2021 Best Paper Honorable Mention Award.
\end{IEEEbiography}
\vspace{-500pt}
\begin{IEEEbiography}[{\includegraphics[width=1in,height=1.25in,clip,keepaspectratio]{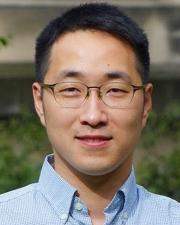}}]{Hyun Soo Park}
Hyun Soo Park is an assistant
professor in the Department of Computer Science and Engineering at the University of Minnesota. He is interested in modeling human and animal behaviors. Prior to joining the UMN, he was a
postdoctoral fellow in the GRASP Lab at the
University of Pennsylvania, and earned his Ph.D.
from Carnegie Mellon University. He received NSF CAREER Award (2019) and CVPR 2021 Best Paper Honorable Mention.
\end{IEEEbiography}








\end{document}